\documentclass[10pt,twocolumn,letterpaper]{article}

\usepackage[pagenumbers]{cvpr} 

\definecolor{cvprblue}{rgb}{0.21,0.49,0.74}
\usepackage[pagebackref,breaklinks,colorlinks,allcolors=cvprblue]{hyperref}
\usepackage[subtle,mathdisplays=normal,wordspacing=normal]{savetrees}

\def\paperID{6176} 
\def\confName{CVPR}
\def\confYear{2026}

\title{OpenFS: Multi-Hand-Capable Fingerspelling Recognition with Implicit Signing-Hand Detection and Frame-Wise Letter-Conditioned Synthesis}

\author{Junuk Cha\\
KAIST
\and
Jihyeon Kim\\
KT
\and
Han-Mu Park\\
KETI
}

\begin{document}
\maketitle
\begin{abstract}
Fingerspelling is a component of sign languages in which words are spelled out letter by letter using specific hand poses.
Automatic fingerspelling recognition plays a crucial role in bridging the communication gap between Deaf and hearing communities, yet it remains challenging due to the signing-hand ambiguity issue, the lack of appropriate training losses, and the out-of-vocabulary (OOV) problem.
Prior fingerspelling recognition methods rely on explicit signing-hand detection, which often leads to recognition failures, and on a connectionist temporal classification (CTC) loss, which exhibits the peaky behavior problem.
To address these issues, we develop OpenFS, an open-source approach for fingerspelling recognition and synthesis.
We propose a multi-hand-capable fingerspelling recognizer that supports both single- and multi-hand inputs and performs implicit signing-hand detection by incorporating a dual-level positional encoding and a signing-hand focus (SF) loss.
The SF loss encourages cross-attention to focus on the signing hand, enabling implicit signing-hand detection during recognition.
Furthermore, without relying on the CTC loss, we introduce a monotonic alignment (MA) loss that enforces the output letter sequence to follow the temporal order of the input pose sequence through cross-attention regularization.
In addition, we propose a frame-wise letter-conditioned generator that synthesizes realistic fingerspelling pose sequences for OOV words. This generator enables the construction of a new synthetic benchmark, called FSNeo.
Through comprehensive experiments, we demonstrate that our approach achieves state-of-the-art performance in recognition and validate the effectiveness of the proposed recognizer and generator.
Codes and data are available in: \href{https://github.com/AIRC-KETI/OpenFS}{https://github.com/AIRC-KETI/OpenFS}.
\end{abstract}    
\section{Introduction}
\label{sec:intro}
Sign language naturally emerged within the Deaf community as a primary means of communication, encompassing a rich system of hand gestures, facial expressions, and body movements. However, it is challenging to create unique gestures for every proper noun or newly coined word. To address this limitation, a supplementary system called \emph{fingerspelling} (FS) was developed, which borrows the structure of spoken language by representing words letter by letter through specific hand poses. Since fingerspelling plays a key role in expressing technical terms, names, and novel words, its accurate recognition is a crucial component of automatic sign language understanding and, ultimately, for bridging the communication gap between Deaf and hearing communities.

\begin{figure}[t]
    \centering
    \includegraphics[width=1\linewidth]{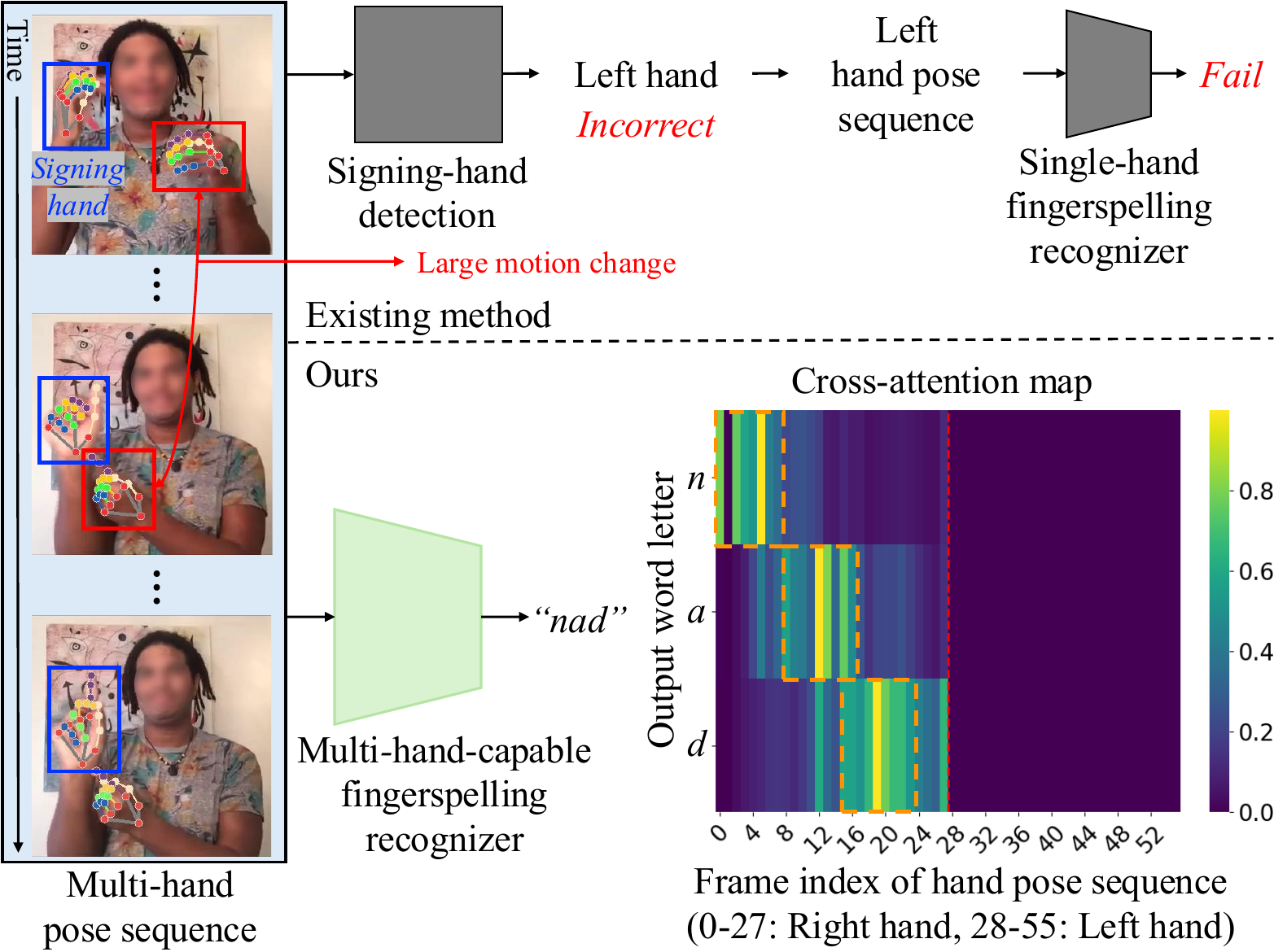}
    \vspace{-7mm}
    \caption{\textbf{Motivation for multi-hand-capable fingerspelling recognition.} The input consists of the hand pose sequence extracted from the video in which the word \emph{``nad"} is fingerspelled using the \emph{right hand}. Existing methods~\cite{shi2018american,shi2019fingerspelling,fayyazsanavi2024fingerspelling} rely on explicit signing-hand detection. However, they misidentify the signing hand when the non-signing hand exhibits large motion changes, which subsequently causes recognition failures. In contrast, our multi-hand-capable fingerspelling recognizer implicitly detects the signing hand from the multi-hand pose sequence to infer the target word. As evidence, the cross-attention map presents high attention values on the frames of the \emph{right hand} when predicting the word \emph{``nad"}.}
    \label{fig:motivation}
    \vspace{-5mm}
\end{figure}

Over the past years, research on fingerspelling recognition has advanced significantly with the adoption of deep learning methods~\cite{shi2018american,shi2019fingerspelling,fayyazsanavi2024fingerspelling}. Shi et al.~\cite{shi2018american} introduced the ChicagoFSWild dataset, and proposed a signing-hand detector and a fingerspelling recognition model based on RGB and optical flow. A larger follow-up dataset, ChicagoFSWildPlus, was later released with an attention-based recognition model that focuses on the signing hand through an optical flow cue~\cite{shi2019fingerspelling}. More recently, PoseNet~\cite{fayyazsanavi2024fingerspelling} employed a Transformer encoder-decoder architecture~\cite{vaswani2017attention} with 2D hand poses from Mediapipe~\cite{lugaresi2019mediapipe}, further enhancing the recognition performance through re-ranking.

Despite recent advances in fingerspelling recognition, existing methods still suffer from three major limitations: the signing-hand ambiguity issue, the peaky behavior problem, and the out-of-vocabulary (OOV) problem. \textbf{1) Signing-hand ambiguity issue}. Most previous methods explicitly detect the signing hand based on optical flow~\cite{shi2018american,shi2019fingerspelling} or the motion change magnitude of hand poses~\cite{fayyazsanavi2024fingerspelling}. However, explicit signing-hand detection is unreliable because the non-signing hand can sometimes exhibit larger movements than the actual signing hand (see Fig.~\ref{fig:motivation}), resulting in unstable training and degraded recognition performance. \textbf{2) Peaky behavior problem.} Existing approaches~\cite{shi2018american,shi2019fingerspelling,fayyazsanavi2024fingerspelling} typically use the CTC loss~\cite{graves2006connectionist}, which often leads the model to predict letters sparsely across frames, a phenomenon known as \emph{peaky behavior}~\cite{zeyer2021does,liu2018connectionist,li2020reinterpreting,huang2024less,yang2023blank,chao2020variational}, thereby providing limited supervision to the encoder and hindering the learning of discriminative hand pose representations. \textbf{3) Out-of-vocabulary problem.} OOV problem in fingerspelling recognition has been largely underestimated. As new vocabulary and neologisms continuously emerge, it is crucial to evaluate whether models can generalize to unseen words and to construct corresponding training data. However, manually collecting data for new words is both labor-intensive and costly, as it requires experts proficient in fingerspelling.

To address these three limitations (\ie, the signing-hand ambiguity issue, the peaky behavior problem, and the OOV problem), we propose OpenFS, an open-source approach for fingerspelling, which comprises three core components for fingerspelling recognition and synthesis:
a multi-hand-capable fingerspelling recognizer,
a monotonic alignment (MA) loss, and
a frame-wise letter-conditioned (FWLC) generator.
\textbf{1) Multi-hand-capable fingerspelling recognizer.}
We introduce a multi-hand-capable fingerspelling recognizer that supports both single- and multi-hand inputs and implicitly identifies the signing hand. It incorporates a dual-level positional encoding and a signing-hand focus (SF) loss. The proposed positional encoding represents both hand identity and temporal position, enabling the model to distinguish between hands while maintaining temporal coherence. The SF loss further encourages the cross-attention to concentrate on the active signing hand.
\textbf{2) Monotonic alignment loss.}
Instead of using the CTC loss, we design a monotonic alignment (MA) loss that enforces monotonic correspondence between the input hand-pose sequence and the output letter sequence. With the dual-level positional encoding, SF loss, and MA loss jointly applied, our recognizer implicitly identifies the signing hand and maintains robust recognition, whereas previous methods~\cite{shi2018american,shi2019fingerspelling,fayyazsanavi2024fingerspelling} relying on explicit signing-hand detection often fail, as shown in Fig.~\ref{fig:motivation}.
\textbf{3) Frame-wise letter-conditioned generator.} We further propose a diffusion-based~\cite{song2020denoising} frame-wise letter-conditioned generator to construct OOV data. Because fingerspelling requires precise modeling of letter-specific articulations over time, conditioning the denoising process on the frame-wise letter sequence enables the generator to learn both fine-grained letter articulations and coherent global transitions across frames. However, existing datasets~\cite{shi2018american,shi2019fingerspelling} lack frame-wise letter annotations, making this training difficult. To overcome this, we introduce a coarse-to-fine frame-wise annotation method that leverages the cross-attention from the trained recognizer and a frame-wise annotation refiner to progressively align pose frames with their corresponding letters. The refined annotations are then used to train the generator, enabling the synthesis of OOV data for scalable training and evaluation.

Furthermore, we introduce a novel synthetic benchmark, FSNeo: FingerSpelling for Neologisms, constructed using our proposed generator to evaluate recognition performance on OOV words. In addition, we synthesize training data consisting of novel words that are not included in the test sets of existing datasets~\cite{shi2018american,shi2019fingerspelling} or FSNeo, which improves the recognition accuracy of both our recognizer and PoseNet~\cite{fayyazsanavi2024fingerspelling}.

Our recognizer is validated on three datasets (ChicagoFSWild~\cite{shi2018american}, ChicagoFSWildPlus~\cite{shi2019fingerspelling}, and FSNeo) for recognition accuracy, while our generator contributes by enabling scalable evaluation and training through the synthesis of OOV data. In addition, our method achieves real-time inference, running significantly faster than the pose-based method~\cite{fayyazsanavi2024fingerspelling}.

To summarize, our main contributions are as follows:
\begin{itemize}
    \item We propose a multi-hand-capable fingerspelling recognizer that leverages a dual-level positional encoding that represents both hand identity and temporal position, along with a signing-hand focus loss and a monotonic alignment loss that jointly enhance cross-attention alignment and improve recognition performance.
    \item We present a coarse-to-fine frame-wise letter annotation method and a frame-wise letter-conditioned generator capable of synthesizing fingerspelling pose sequences for out-of-vocabulary words. Using the generator, we construct a new synthetic benchmark, FSNeo.
    \item Our proposed recognizer achieves state-of-the-art recognition performance on ChicagoFSWild, ChicagoFSWildPlus, and FSNeo, with more than 100 times faster inference than the existing pose-based method, without any post-processing.
\end{itemize}

\section{Related Work}
\label{sec:related_work}

\noindent\textbf{Fingerspelling recognition.}
Recent advances in computer vision have led to the development of a variety of sign language recognition methods based solely on RGB visual input~\cite{koller2017re,koller2016deep,koller2016deep_bmvc,koller2018deep,pannattee2024american,shi2021fingerspelling,shi2017multitask,cui2017recurrent,huang2018video,shi2018american,cihan2017subunets,kumwilaisak2022american,shi2019fingerspelling,pu2019iterative,pannattee2021novel,kabade2023american,shi2022searching,li2022multi,gajurel2021fine,li2023contrastive}.
For fingerspelling recognition,
Shi~\etal~\cite{shi2018american} collected in-the-wild videos and manually annotated them.
In this work, they also proposed a method that incorporates a hand detector along with CNN and LSTM architectures to build a fingerspelling recognizer.
Subsequently, Shi~\etal~\cite{shi2019fingerspelling} collected a larger set of in-the-wild video data and improved recognition performance by leveraging both the increased data scale and an iterative visual attention mechanism.
Gajurel~\etal~\cite{gajurel2021fine} proposed a Transformer-based model with fine-grained visual attention and a training strategy using CTC~\cite{graves2006connectionist} loss and maximum-entropy~\cite{dubey2018maximum} loss to improve fingerspelling recognition in in-the-wild video sequences.
FSS-Net~\cite{shi2022searching} demonstrated the importance of fingerspelling detection as a key component of a search and retrieval model in real world scenario.
However, RGB-based methods suffer from significant limitations in data scalability and domain generalizability.
Therefore, pose-based methods~\cite{uthus2023youtube,fayyazsanavi2024fingerspelling,tunga2021pose,bohavcek2022sign,naz2023signgraph,song2025hand,papadimitriou2020multimodal,parelli2020exploiting,jiang2021skeleton} have emerged as a promising alternative that effectively mitigates the domain gap.
PoseNet~\cite{fayyazsanavi2024fingerspelling} proposed Transformer~\cite{vaswani2017attention} encoder-decoder-based model that takes a sequence of single hand pose as input and predicts letters using a re-ranking at inference. The re-ranking step involves fusing encoder features and decoder features to improve final letter prediction.
HandReader~\cite{korotaev2025handreader} introduced a multi-modal framework with a temporal shift-adaptive module (RGB) and a temporal pose encoder (pose).

However, in fingerspelling recognition, the issues of unstable signing-hand detection and the peaky behavior~\cite{zeyer2021does} caused by CTC loss~\cite{graves2006connectionist} have not been fully explored. To address these challenges, we propose a multi-hand-capable recognizer incorporating a dual-level positional encoding, a signing-hand focus loss, and a monotonic alignment loss.

\noindent\textbf{Fingerspelling generation.}
To address the out-of-vocabulary (OOV) problem in fingerspelling recognition, generating fingerspelling pose sequences is essential.
Research on human motion generation has focused on body movements conditioned on text~\cite{ahuja2019language2pose,athanasiou2022teach,delmas2022posescript,guo2022generating,petrovich2021action,plappert2018learning,zhang2024motiondiffuse,zhao2023modiff,zhang2023generating,tevet2023human,guo2024momask,tevet2022motionclip,lin2023being,meng2025rethinking}.
Hand motion has also been studied in domains such as hand-object interaction~\cite{cha2024text2hoi,ghosh2023imos,zhang2021manipnet,zheng2023cams,christen2024diffh2o,huang2025hoigpt,li2025latenthoi}, focusing on physical plausibility.
Sign language generation has been explored~\cite{saunders2020progressive,saunders2021mixed,stoll2022there,baltatzis2024neural,yu2024signavatars,bensabath2025text}, which is modeled as full-body motion but places particular emphasis on hand articulation for semantic expressiveness.
However, generating fingerspelling with existing text-to-motion models~\cite{guo2024momask,tevet2022motionclip,lin2023being,tevet2023human,zhang2024motiondiffuse,meng2025rethinking}, which typically rely on the CLIP encoder~\cite{radford2021learning}, results in locally smoothed motions and poor ordering consistency. 
Because these models capture word-level semantics and use them as global conditioning, they are inadequate for fingerspelling, which requires letter-level conditioning where each letter corresponds to a specific hand pose and the transitions between poses must be modeled.
This limitation highlights the need for approaches specifically designed for the unique characteristics of fingerspelling.
Therefore, we propose a frame-wise letter-conditioned generator that synthesizes fingerspelling sequences with locally realistic and pose-accurate hand articulations.
\section{Method}
\label{sec:method}

\begin{figure}[t]
    \centering
    \includegraphics[width=1\linewidth]{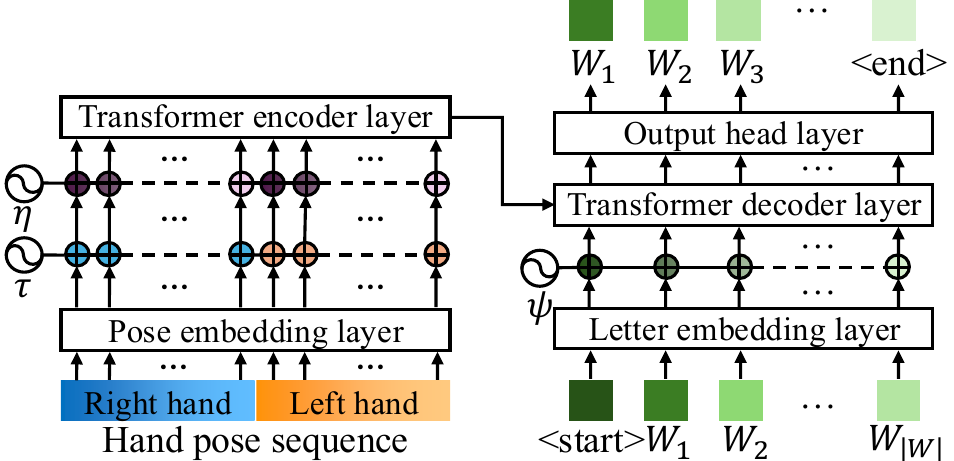}
    \vspace{-7mm}
    \caption{\textbf{Overview of the multi-hand-capable fingerspelling recognizer.}
    The hand pose sequence is embedded into a feature space and encoded using our proposed dual-level positional encoding, which consists of hand-identity encoding ($\tau$) and temporal positional encoding ($\eta$). The recognizer's decoder then predicts the next letter token based on the pose-aware, semantically rich encoder features. $\psi$ denotes the standard positional encoding~\cite{vaswani2017attention}, and $W_i$ represents the $i$-th letter of the word. \ \textless start\textgreater{} and \textless end\textgreater{} are special tokens indicating the start and end of the letter token sequence, respectively.
    }
    \vspace{-5mm}
    \label{fig:pipeline_recognizer}
\end{figure}

In this section, we present OpenFS: multi-hand-capable fingerspelling recognizer and frame-wise letter-conditioned generator, along with a new synthetic benchmark called FSNeo: FingerSpelling for Neologisms.

We propose a multi-hand-capable fingerspelling recognizer equipped with a dual-level positional encoding, a signing-hand focus loss, and a monotonic alignment loss, eliminating errors caused by explicit signing-hand detection and avoiding reliance on the CTC loss (Sec.~\ref{sec:method_recognizer}). For the OOV problem, generating fingerspelling pose sequences is essential. Thus, we propose both a coarse-to-fine frame-wise letter annotation method and a frame-wise letter-conditioned generator (Sec.~\ref{sec:method_generator}). Furthermore, we construct a novel benchmark, FSNeo, to evaluate the recognition models on neologisms, employing our frame-wise letter-conditioned generator (Sec.~\ref{sec:method_novel_benchmark}).

\begin{figure*}[t]
    \centering
    \includegraphics[width=1\linewidth]{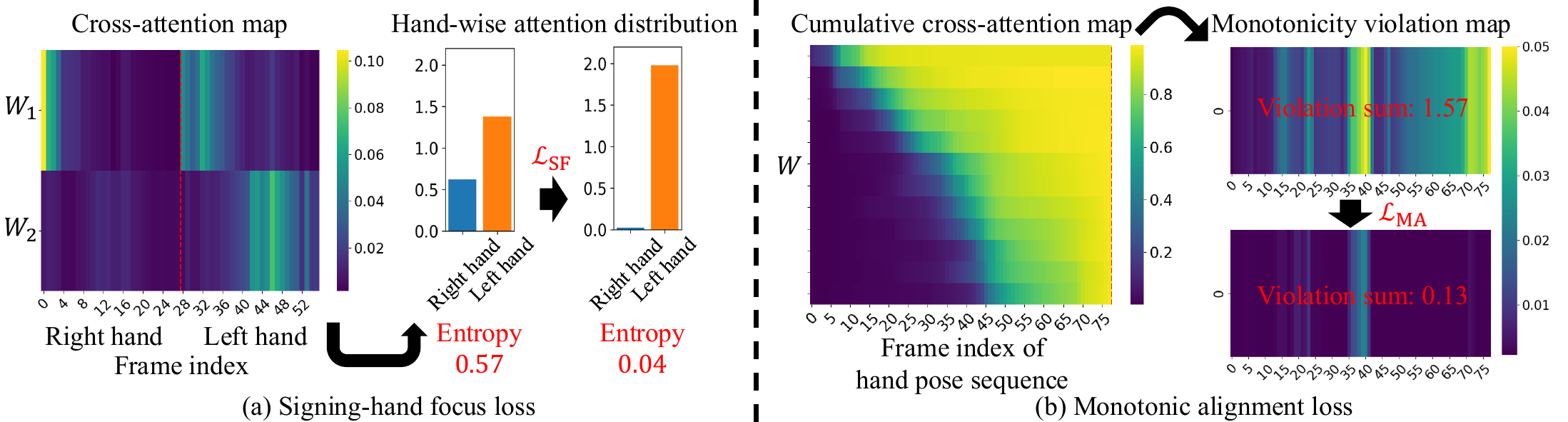}
    \vspace{-7mm}
    \caption{\textbf{Overview of the signing-hand (SF) and monotonic alignment (MA) losses.}
    (a) The signing-hand focus (SF) loss $\cal{L}_\text{SF}$ measures the entropy of the hand-wise attention distribution derived from the cross-attention map between input hand pose tokens and output letter tokens. Minimizing this entropy encourages the recognizer to focus on the single signing hand.
    (b) The monotonic alignment (MA) loss $\cal{L}_\text{MA}$ penalizes misalignments that violate the natural temporal order between input hand pose tokens and output letter tokens in fingerspelling. Reducing these violations encourages the model to interpret the hand pose tokens in a temporally coherent manner to predict the letter token.
    }
    \label{fig:losses}
    \vspace{-5mm}
\end{figure*}

\subsection{Multi-Hand-Capable Fingerspelling Recognizer}
\label{sec:method_recognizer}
An overview of our multi-hand-capable fingerspelling recognizer is illustrated in Fig.~\ref{fig:pipeline_recognizer}. We adopt a Transformer encoder-decoder architecture~\cite{vaswani2017attention} for fingerspelling recognition. The encoder takes a normalized 2D single- or multi-hand pose sequence extracted using an off-the-shelf pose estimator~\cite{lugaresi2019mediapipe} and embeds it with an MLP layer. The resulting embeddings are then encoded with a dual-level positional encoding, consisting of hand-identity and temporal positional components, enabling the encoder to process the pose sequence. The decoder takes as input a letter sequence derived from the target word, augmented with special start and end tokens. This sequence is embedded using a token embedding layer and standard positional encoding~\cite{vaswani2017attention}. Finally, the decoder attends to both the encoded pose features and its own inputs to predict the next letter token in the sequence.

\noindent\textbf{Dual-level positional encoding.} Unlike the standard positional encoding~\cite{vaswani2017attention}, which assigns a single positional index to each token, we design a dual-level encoding scheme that separately encodes 1) the hand identity, including both the hand side (right/left) and the person identity, and 2) the temporal position of each pose frame. We use a sinusoidal formulation following~\cite{vaswani2017attention} for both hand identity encoding and temporal positional encoding. The same hand identity encoding is shared across all tokens belonging to the same hand to distinguish different hand identities, while the same temporal positional encoding value is shared by multiple hands at the same frame index to maintain temporal alignment and distinct values are used across frames to preserve temporal ordering. These encoding values are added to the pose token embeddings and then fed into the Transformer encoder layers.

\begin{figure*}
    \centering
    \includegraphics[width=1\linewidth]{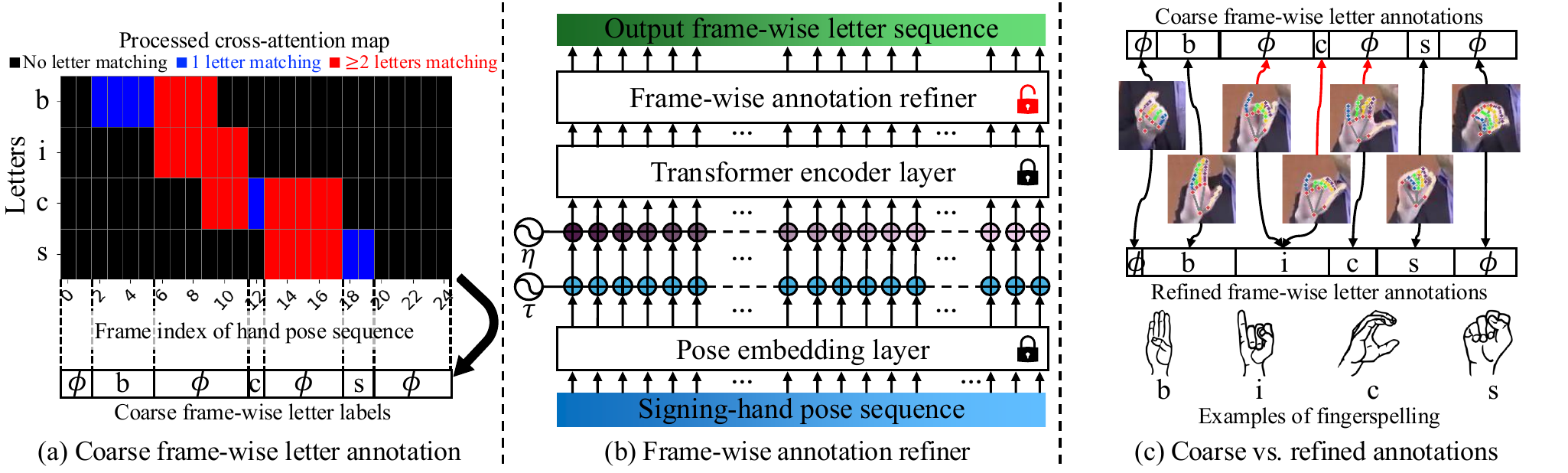}
    \vspace{-7mm}
    \caption{\textbf{Overview of the coarse-to-fine frame-wise letter annotation method.}
    (a) We utilize cross-attention map between input hand pose tokens and output letter tokens to generate coarse frame-wise letter annotations, where $\phi$ denotes a non-letter annotation.
    (b) To refine the coarse frame-wise letter annotations, we freeze the pre-trained recognizer and train a frame-wise annotation refiner supervised by the coarse frame-wise letter annotations.
    (c) The trained frame-wise annotation refiner produces refined frame-wise letter annotations. The coarse and refined annotations are compared with the corresponding image frames, where each label–frame pair is linked with arrows, and mismatched cases are highlighted in red.}
    \label{fig:pipeline_annotation}
    \vspace{-3mm}
\end{figure*}

\begin{figure}
    \centering
    \includegraphics[width=1\linewidth]{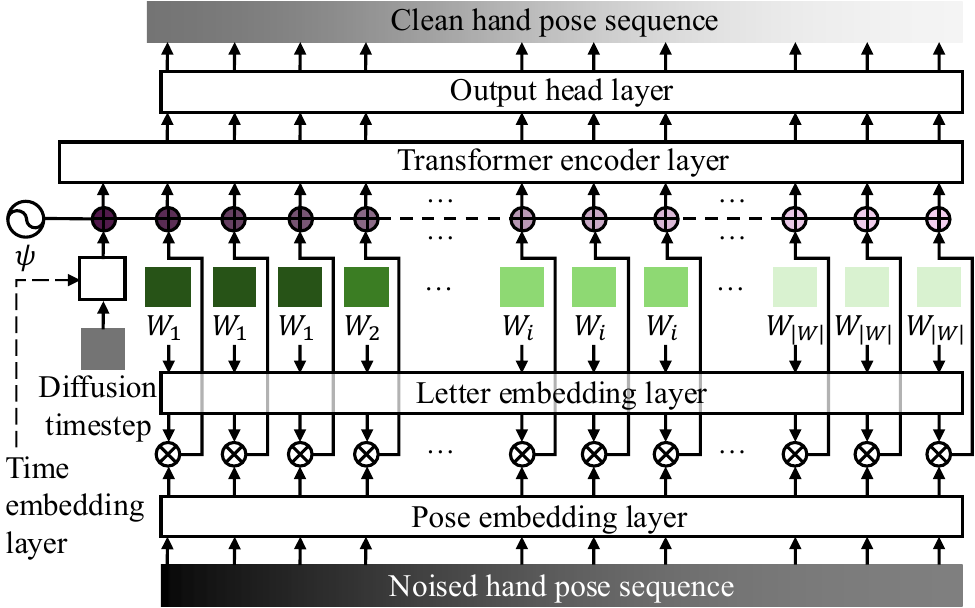}
    \vspace{-7mm}
    \caption{\textbf{Overview of the frame-wise letter-conditioned generator.} $W_i$ is the $i$-th letter of the word, $|W|$ is the word length, $\otimes$ denotes concatenation, and $\psi$ denotes the standard positional encoding~\cite{vaswani2017attention}. The generator embeds each letter token and each noised pose vector through their respective embedding layers. The resulting letter and pose embeddings are concatenated frame-wise and, given a diffusion timestep, are denoised by the generator encoder to produce a clean hand-pose sequence.}
    \label{fig:pipeline_generator}
    \vspace{-5mm}
\end{figure}

\noindent\textbf{Signing-Hand Focus Loss and Monotonic Alignment Loss.} Along with a cross-entropy loss $\cal{L}_\text{CE}$ applied to the decoder outputs, we propose auxiliary loss function $\mathcal{L}_\text{aux}$ consisting of signing-hand focus (SF) loss $\mathcal{L}_\text{SF}$ and monotonic alignment (MA) loss $\mathcal{L}_\text{MA}$:
\begin{eqnarray}
    \mathcal{L} &=& \mathcal{L}_\text{CE} + \mathcal{L}_\text{aux}, \\
    \mathcal{L}_\text{aux} &=& \lambda_\text{SF} \mathcal{L}_\text{SF} + \lambda_\text{MA} \mathcal{L}_\text{MA}.
    \vspace{-3mm}
\end{eqnarray}
The SF loss $\mathcal{L}_\text{SF}$ encourages the decoder to focus the dominant signing hand by minimizing the entropy of the hand-wise attention distribution, computed from the cross-attention.
The MA loss $\mathcal{L}_\text{MA}$ guides the cross-attention to follow a monotonic alignment, reflecting the fact that the temporal order of the fingerspelling pose sequence is aligned with the sequential order of the letters. Fig.~\ref{fig:losses} presents the overview and effectiveness of the proposed losses. $\lambda_\text{SF}$ and $\lambda_\text{MA}$ denote the corresponding weights, which are empirically set to 0.8 and 1.0, respectively.

To compute the \textbf{signing-hand focus loss} $\mathcal{L}_\text{SF}$, we utilize the decoder’s cross-attention weights between input hand pose tokens and output letter tokens. The cross-attention scores are averaged across decoder layers to form a layer-averaged attention map. Each pose token is labeled according to its hand identity, and the attention contributions are aggregated per hand to obtain the hand-wise attention distribution for every letter token, as shown in Fig.~\ref{fig:losses}(a). The entropy of this distribution reflects the model’s uncertainty in determining which hand contributes to the letter tokens. By minimizing this entropy, the SF loss $\mathcal{L}_\text{SF}$ encourages the decoder to better identify the dominant hand and focus its attention primarily on the correct signing hand to achieve better fingerspelling recognition performance.

To compute the \textbf{monotonic alignment loss} $\mathcal{L}_\text{MA}$, a cumulative cross-attention map is first constructed to track how the attention accumulates over time across letter tokens. To measure the temporal change of attention between consecutive letters, we compute the difference of these cumulative maps along the letter dimension. Positive values in this difference indicate cases where a later letter assigns more attention to earlier frames than its preceding letter, violating the natural temporal order of fingerspelling and causing confusion in recognition. Such positive deviations are regarded as monotonicity violations and are penalized through the MA loss $\mathcal{L}_\text{MA}$, encouraging smooth and monotonic attention transitions across the decoded letter sequence, as shown in Fig.~\ref{fig:losses}(b). More detailed loss formulations are provided in the supplementary material.

\subsection{Frame-Wise Letter-Conditioned Generator}
\label{sec:method_generator}
Our generator is built upon a Transformer encoder architecture~\cite{vaswani2017attention} combined with a diffusion mechanism~\cite{song2020denoising}, where the diffusion process enhances motion fidelity and expressiveness through iterative refinement. The generator takes as input a noised hand-pose sequence and a frame-wise letter sequence, the latter produced by our coarse-to-fine frame-wise letter annotation method. Each sequence is independently transformed into pose and letter embeddings through their respective embedding layers. The resulting embeddings are concatenated frame-wise, added with standard positional encoding~\cite{vaswani2017attention}, and passed to the Transformer encoder together with diffusion timestep embeddings. The encoder learns to map these inputs to clean poses that capture the fine-grained structure of fingerspelling. The generator is trained using a mean squared error (MSE) loss~\cite{tevet2023human} between the predicted clean pose sequence and the ground-truth sequence, enabling precise modeling of pose–letter relationships and smooth temporal transitions. An overview of the coarse-to-fine annotation method is shown in Fig.~\ref{fig:pipeline_annotation}, and the overall frame-wise letter-conditioned generator is illustrated in Fig.~\ref{fig:pipeline_generator}.

\noindent\textbf{Coarse frame-wise letter annotation.}
Existing datasets provide RGB video frames paired with word annotations but lack frame-wise letter annotations. To obtain such annotations, we leverage the recognizer’s cross-attention matrix, which captures the alignment between output letter tokens and input pose frames. Specifically, we compute the layer-averaged cross-attention matrix, where each entry indicates how strongly an output letter token attends to particular pose frames. Frames with attention weights exceeding a threshold are assigned to the corresponding output letter, but, if multiple letters are assigned to a frame or if no letter is assigned, the frame is labeled as blank ($\phi$), as shown in Fig.~\ref{fig:pipeline_annotation}(a). These frame assignments are then used as coarse frame-wise letter annotations. More detailed formulations and explanations are provided in the supplementary material.

\noindent\textbf{Refined frame-wise letter annotation.}
However, these coarse frame-wise letter annotations are inherently noisy, as the recognizer is not explicitly trained with a frame-wise classification objective. To refine them, we freeze the pre-trained recognizer and train a frame-wise annotation refiner that takes the encoder features as input and predicts a letter for each frame, as illustrated in Fig.~\ref{fig:pipeline_annotation}(b). The annotation refiner is supervised with the coarse frame-wise letter annotations using the cross-entropy loss $\mathcal{L}_\text{CE}$, where a lower weight of 0.1 is assigned to the blank ($\phi$) class to mitigate its dominance. The refiner is trained in a fully data-driven manner without any additional heuristics. After training, it produces refined frame-wise letter annotations, which are subsequently used as supervision for training the hand pose generator. As shown in Fig.~\ref{fig:pipeline_annotation}(c), the refined annotations provide more accurate and consistent frame-wise annotations compared to the coarse annotations.

\noindent\textbf{Fingerspelling pose generation.}
At inference, given a letter sequence as input, the model iteratively denoises a noised pose sequence over multiple steps (\eg, 50 iterations)~\cite{song2020denoising}. At each diffusion timestep, it predicts a clean pose sequence from the current noisy sample and reconstructs a slightly less noisy sample for the previous diffusion timestep by partially reapplying noise, following a sampling strategy similar to MDM~\cite{tevet2023human}. This iterative denoising continues until the diffusion timestep reaches 0, yielding a fully refined pose sequence that is both natural in dynamics and faithful to letter-level finger articulations.

\subsection{Novel Benchmark for OOV Evaluation}
\label{sec:method_novel_benchmark}
We construct a novel benchmark, FSNeo (FingerSpelling for Neologisms), to evaluate recognition performance on neologisms. To this end, we employ our frame-wise letter-conditioned generator and adopt the terminology from NEO-BENCH~\cite{zheng2024neo}. NEO-BENCH defines three categories of neologisms: 1) lexical neologisms, words denoting newly emerging concepts; 2) morphological neologisms, blends derived from existing subwords; and 3) semantic neologisms, pre-existing words that acquire new meanings. Following this taxonomy, FSNeo comprises 1,635 unique words, and for each word, five pose sequences are generated to enhance diversity.
Each entry in FSNeo consists of a word–pose pair, where the pose is represented as a 3D hand pose sequence. In total, FSNeo contains 8,175 samples and serves as a benchmark for evaluating out-of-vocabulary (OOV) fingerspelling recognition.

We further describe the data generation process as follows: an arbitrary word is first converted into a letter sequence, which serves as the input to the frame-wise letter-conditioned generator. Since the hand pose corresponding to each letter persists over time, each letter naturally spans multiple frames. To simulate this temporal duration, the number of repetitions per letter is randomly chosen between 3 and 10, while space characters are repeated 2 or 3 times, reflecting the observed data distribution~\cite{shi2018american,shi2019fingerspelling}. The resulting frame-level letter sequence is then fed into the generator to produce diverse pose sequences.
\section{Experiments}
\label{sec:experiments}

\subsection{Dataset}
The Chicago fingerspelling dataset~\cite{shi2018american,shi2019fingerspelling} comprises videos of individuals performing American Sign Language (ASL) fingerspelling, gathered from online sources. ChicagoFSWild~\cite{shi2018american} includes 7,304 sequences from 160 signers, while its extended version, ChicagoFSWild+~\cite{shi2019fingerspelling}, contains 55,232 sequences from 260 signers. Although both datasets provide word annotations, they do not provide frame-level letter annotations. We follow the train/dev/test split introduced in the Chicago fingerspelling dataset~\cite{shi2018american,shi2019fingerspelling}. We also utilize FSNeo for evaluation, and the details are described in Sec.~\ref{sec:method_novel_benchmark}. To evaluate signing-hand detection accuracy, we manually annotate the signing hand for each sample in CFSW~\cite{shi2018american}, specifying whether the right or left hand is being used. These manual annotations will be released for future research.

For an additional training dataset, we synthesize fingerspelling pose sequences using terminology from the English words dataset~\footnote{https://www.kaggle.com/datasets/bwandowando/479k-english-words}. We exclude words that appear in the test set and follow the same procedure described in Sec.~\ref{sec:method_novel_benchmark}. Our model is trained on CFSW~\cite{shi2018american} and CFSWP~\cite{shi2019fingerspelling}, and results marked with the symbol $\dagger$ indicate that additional synthesized training data are included.

\subsection{Competing Methods}
For fair comparison, we report approaches~\cite{shi2018american,shi2019fingerspelling,shi2022searching,fayyazsanavi2024fingerspelling} that are trained on CFSW~\cite{shi2018american} and CFSWP~\cite{shi2019fingerspelling}, and utilize the publicly released implementation and checkpoints of PoseNet~\cite{fayyazsanavi2024fingerspelling}. RGB image-based methods~\cite{shi2018american,shi2019fingerspelling,shi2022searching} directly predict fingerspelling letters from video frames. PoseNet~\cite{fayyazsanavi2024fingerspelling}, in contrast, employs an off-the-shelf hand pose estimator~\cite{lugaresi2019mediapipe} to obtain hand pose inputs and then predicts the corresponding fingerspelling letters. We cannot evaluate RGB image-based methods~\cite{shi2018american,shi2019fingerspelling,shi2022searching} on FSNeo because RGB images are not available.
\subsection{Evaluation Metric}
For recognition evaluation, following the previous works~\cite{shi2018american,shi2019fingerspelling,shi2022searching,fayyazsanavi2024fingerspelling}, we adopt letter accuracy (Acc.), defined as $1-\frac{S + D + I}{N}$, where $S$, $D$, and $I$ denote the number of substitutions, deletions, and insertions, respectively, and $N$ is the number of letters. In addition, we add \textit{Fair Acc.}, which measures letter accuracy only on samples where PoseNet~\cite{fayyazsanavi2024fingerspelling} successfully detects the signing hand. This metric removes variance introduced by signing-hand detection failures and focuses purely on recognition capability. We also report the \textit{IV Acc.} and the \textit{OOV Acc.}, which denote the letter accuracy on in-vocabulary and out-of-vocabulary words, respectively. These metrics assess both recognition performance on seen words and the generalization ability to unseen words during training.

For the speed evaluation, we report four metrics: latency ($t_{lat}$), throughput ($R_{tp}$), letters per second ($R_{lps}$), and pose frames per second (FPS). Latency denotes the total inference time, in seconds, required to process all 868 samples~\cite{shi2018american}. Throughput measures the number of samples processed per second, while letters per second quantifies the number of recognized letters divided by the total inference time. FPS measures the number of pose frames processed per second. A lower latency indicates faster processing, whereas higher throughput, letters per second, and FPS correspond to greater efficiency.

\begin{table}[t]
    \centering
    \caption{We evaluate our model on CFSW~\cite{shi2018american}, CFSWP~\cite{shi2019fingerspelling} and FSNeo, in terms of letter accuracy, comparing it with both RGB-based approaches~\cite{shi2018american,shi2019fingerspelling,shi2022searching} and the pose-based approach, PoseNet~\cite{fayyazsanavi2024fingerspelling}. Parenthesized numbers indicate the accuracy improvements achieved by using the additional synthetic data$^\dagger$.}
    \vspace{-2mm}
    \setlength{\tabcolsep}{4pt}
    \begin{tabular}{l|ccc}
    \hline
    Method & CFSW~\cite{shi2018american} & CFSWP~\cite{shi2019fingerspelling} & FSNeo \\
    \hline
    Shi~\etal~\cite{shi2018american} & 57.5 & 58.3 & - \\
    Shi~\etal~\cite{shi2019fingerspelling} & 61.2 & 62.3 & - \\
    FSS-Net~\cite{shi2022searching} & 52.5 & 64.4 & - \\
    \hline
    PoseNet~\cite{fayyazsanavi2024fingerspelling} & 61.6 & 61.0 & 61.2 \\
    Ours & \textbf{75.4} & \textbf{70.5} & \textbf{80.5} \\
    \hline
    PoseNet$^\dagger$~\cite{fayyazsanavi2024fingerspelling} & 69.2(\textbf{+7.6}) & 69.4(\textbf{+8.4}) & 94.9(\textbf{+33.7}) \\
    Ours$^\dagger$ & \textbf{77.7}(+2.3) & \textbf{74.6}(+4.1) & \textbf{97.6}(+17.1) \\
    \hline
    \end{tabular}
    \label{tab:quantitative_comparison}
\end{table}

\begin{table}[t]
    \centering
    \caption{On the ChicagoFSWild~\cite{shi2018american} dataset, Fair Acc. denotes letter accuracy on samples where PoseNet~\cite{fayyazsanavi2024fingerspelling} successfully detects the signing hand. IV Acc. denotes letter accuracy on in-word samples, while OOV Acc. denotes letter accuracy on out-of-vocabulary samples. Parenthesized numbers indicate the accuracy improvements achieved by using the additional synthetic data$^\dagger$.}
    \vspace{-2mm}
    \begin{tabular}{l|ccc}
    \hline
    Method & Fair Acc. & IV Acc. & OOV Acc. \\
    \hline
    PoseNet~\cite{fayyazsanavi2024fingerspelling} & 66.2 & 65.0 & 58.8 \\
    Ours & \textbf{76.0} & \textbf{80.1} & \textbf{71.8} \\
    \hline
    PoseNet$^\dagger$~\cite{fayyazsanavi2024fingerspelling} & 74.5(\textbf{+8.3}) & 70.9(\textbf{+5.9}) & 67.9(\textbf{+9.1}) \\
    Ours$^\dagger$ & \textbf{77.9}(+1.9) & \textbf{84.5}(+4.4) & \textbf{72.5}(+0.7) \\
    \hline
    \end{tabular}
    \label{tab:quantitative_comparison_fair}
\end{table}

\begin{table}[t]
    \centering
    \caption{Comparison of signing-hand detection accuracy on the ChicagoFSWild~\cite{shi2018american}. The symbol $\dagger$ denotes methods trained with additional synthetic data.}
    \vspace{-2mm}
    \begin{tabular}{c|c|c|c}
    \hline
    Method & Accuracy & Method & Accuracy \\
    \hline
    PoseNet~\cite{fayyazsanavi2024fingerspelling} & 90.4 & Ours/Ours$^\dagger$ & \textbf{99.9} \\
    \hline
    \end{tabular}
    \label{tab:hand_detection}
\end{table}

\subsection{Comparison}
\noindent\textbf{Quantitative comparison.} 
From Tab.~\ref{tab:quantitative_comparison}, we observe that our model achieves consistent improvements across all benchmarks. On \textit{ChicagoFSWild} (CFSW)~\cite{shi2018american} and \textit{ChicagoFSWildPlus} (CFSWP)~\cite{shi2019fingerspelling}, our approach outperforms prior methods, including Shi et al.~\cite{shi2018american,shi2019fingerspelling}, FSS-Net~\cite{shi2022searching}, and the pose-based approach, PoseNet~\cite{fayyazsanavi2024fingerspelling}. The performance gain in these challenging in-the-wild settings demonstrates the robustness of our model to diverse fingerspelling styles across different signers. On \textit{FSNeo}, which is designed to evaluate generalization to neologism fingerspelling, our method shows a clear advantage. These results highlight the strong generalization ability of our model to out-of-vocabulary cases, where previous method~\cite{fayyazsanavi2024fingerspelling} performs considerably worse. In Tab.~\ref{tab:quantitative_comparison_fair}, \textit{Fair Acc.} shows that our model outperforms PoseNet~\cite{fayyazsanavi2024fingerspelling} under the same successful detection scenario in the ChicagoFSWild~\cite{shi2018american} dataset. Furthermore, it achieves higher scores in both \textit{IV Acc.} (in-distribution words) and \textit{OOV Acc.} (out-of-distribution words), showing better generalization to both unseen signers and unseen words. In Tab.~\ref{tab:hand_detection}, we compare our implicit signing-hand detection performance with the explicit pose-based approach~\cite{fayyazsanavi2024fingerspelling}. We leverage cross-attention to detect the signing hand. Our detection outperforms the explicit pose-based approach on ChicagoFSWild~\cite{shi2018american}. Notably, our method fails in only a single case, which is also ambiguous even for humans without prior context (\eg, the intended word), since the signing hand appears only in the later part of a short video. A detailed example can be found in the supplementary material.

Moreover, as shown in~\cref{tab:quantitative_comparison,tab:quantitative_comparison_fair}, the synthetic data generated by our frame-wise letter-conditioned generator not only improves the performance of our recognition model but also significantly enhances the accuracy of PoseNet~\cite{fayyazsanavi2024fingerspelling}, further validating the effectiveness and versatility of our data generation approach.

\begin{figure}
    \centering
    \includegraphics[width=0.95\linewidth]{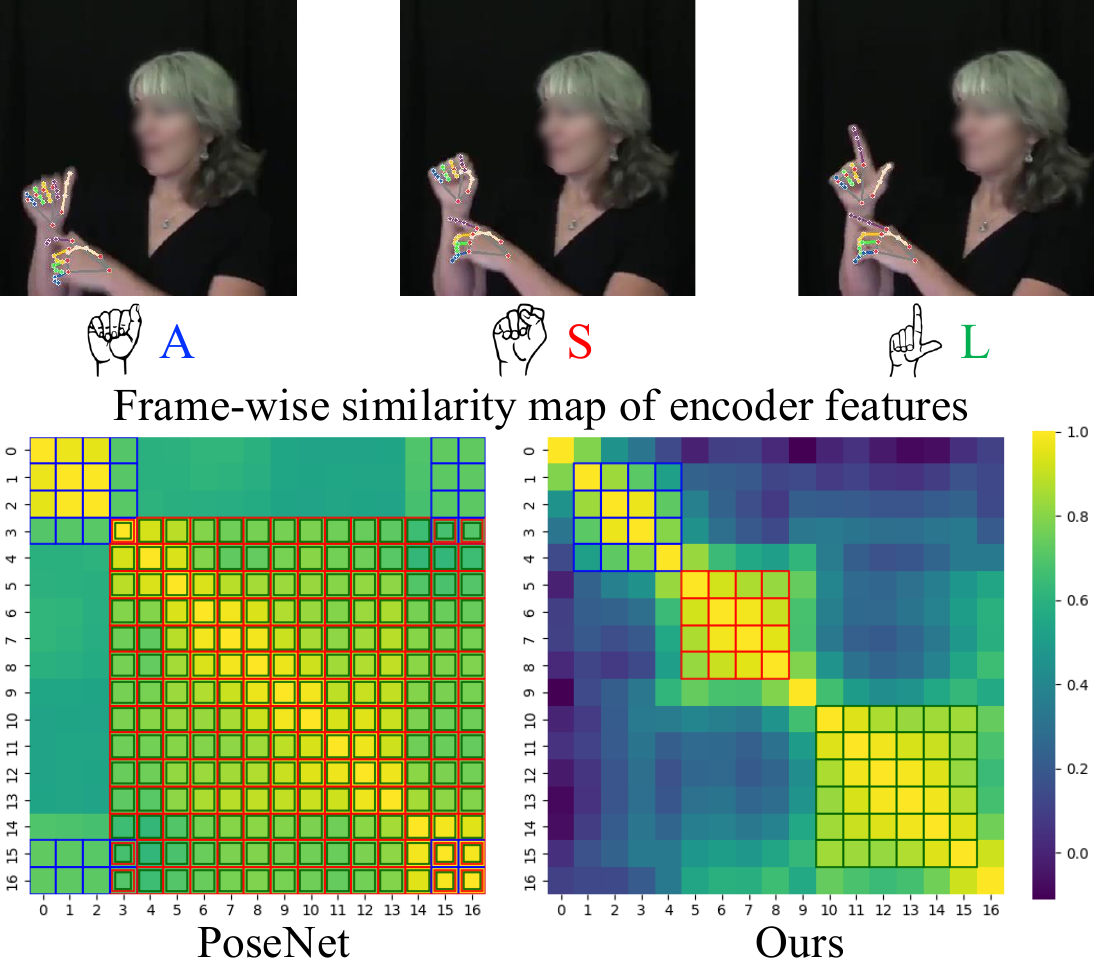}
    \vspace{-3mm}
    \caption{Top: image frames corresponding to the letters ``A'', ``S'', and ``L''. Bottom: frame-wise similarity maps of encoder features for PoseNet~\cite{fayyazsanavi2024fingerspelling} and our recognizer on the ``ASL'' sample, presenting how similarly each encoder represents hand poses across frames. Blue, red, and green borders indicate frames with similar features to the letters ``A'', ``S'', and ``L'', respectively.}
    \label{fig:redundant_features}
\end{figure}

\noindent\textbf{Qualitative comparison.} In Fig.~\ref{fig:redundant_features}, we visualize the frame-wise similarity maps of encoder features from PoseNet~\cite{fayyazsanavi2024fingerspelling} and our recognizer to compare how each model encodes temporal relationships between hand poses. In the similarity maps, blue, red, and green denote feature regions corresponding to the letters ``A", ``S", and ``L", respectively. PoseNet, trained with the CTC loss~\cite{graves2006connectionist}, often exhibits peaky behavior~\cite{zeyer2021does}, producing sparse letter predictions across frames and resulting in semantically weak representations. Consequently, PoseNet struggles to distinguish pose similarities and differences, predicting only ``A" for the ground-truth word ``ASL". In contrast, our recognizer produces semantically rich representations by capturing similarities among alike hand poses and emphasizing differences among distinct ones, thereby correctly predicting the word ``ASL".

Additional experiments and results, including comparisons of input pose representations, conditioning strategies, model efficiency, error-type sensitivity, robustness to long words, additional metrics, evaluation on FSBoard~\cite{georg2025fsboard}, and more qualitative results, are provided in the supplementary material.

\begin{table}[t]
    \centering
    \caption{Ablation study on the effects of positional encoding (pos. enc.) and auxiliary loss ($\mathcal{L}_\text{aux}$) on letter accuracy. “w” and “w/o” denote “with” and “without,” respectively. The standard pos. enc. refers to the standard sinusoidal encoding~\cite{vaswani2017attention}, while the dual-level pos. enc. represents our proposed encoding that captures both temporal and hand identity information. The last row corresponds to our full model (Ours).}
    \vspace{-2mm}
    \begin{tabular}{l|c}
    \hline
    Method & CFSW~\cite{shi2018american} \\
    \hline
    w standard pos. enc. \& w/o $\cal{L}_\text{aux}$ & 73.2 \\
    w standard pos. enc. \& w $\cal{L}_\text{aux}$ & 73.1 \\
    w dual-level pos. enc. \& w/o $\cal{L}_\text{aux}$ & 74.8 \\
    w dual-level pos. enc. \& w $\cal{L}_\text{aux}$ & \textbf{75.4} \\
    \hline
    \end{tabular}
    \label{tab:ablation_PE_loss}
\end{table}

\subsection{Ablation Study}
\noindent\textbf{Recognizer.}
In Tab.~\ref{tab:ablation_PE_loss}, we conduct an ablation study on the dual-level positional encoding and auxiliary losses, signing-hand focus loss and monotonic alignment loss. Both components show clear benefits: adopting the dual-level positional encoding and adding the auxiliary losses improve overall performance. Notably, the auxiliary losses exhibit a synergistic effect only when used together with the dual-level positional encoding, highlighting the importance of the proposed encoding scheme.

\noindent\textbf{Generator.}
We compare three types of conditioning signals:WC (word-conditioned), LC (letter-conditioned), and FWLC (frame-wise letter-conditioned). In the WC setting, CLIP~\cite{radford2021learning} is employed to extract textual features from the entire input word, and the generator conditions on these word-level features to provide coarse-grained semantic guidance for the whole sequence. In the LC setting, each letter of the input word is individually embedded and treated as an independent token, forming a global representation that captures the overall spelling context to guide the motion sequence. In the FWLC setting, each frame is conditioned on its corresponding letter token, enabling fine-grained and temporally aligned control. The local conditioning leverages a stronger prior at the frame level, resulting in more accurate and sharper motion generation.

\begin{figure}
    \centering
    \includegraphics[width=1\linewidth]{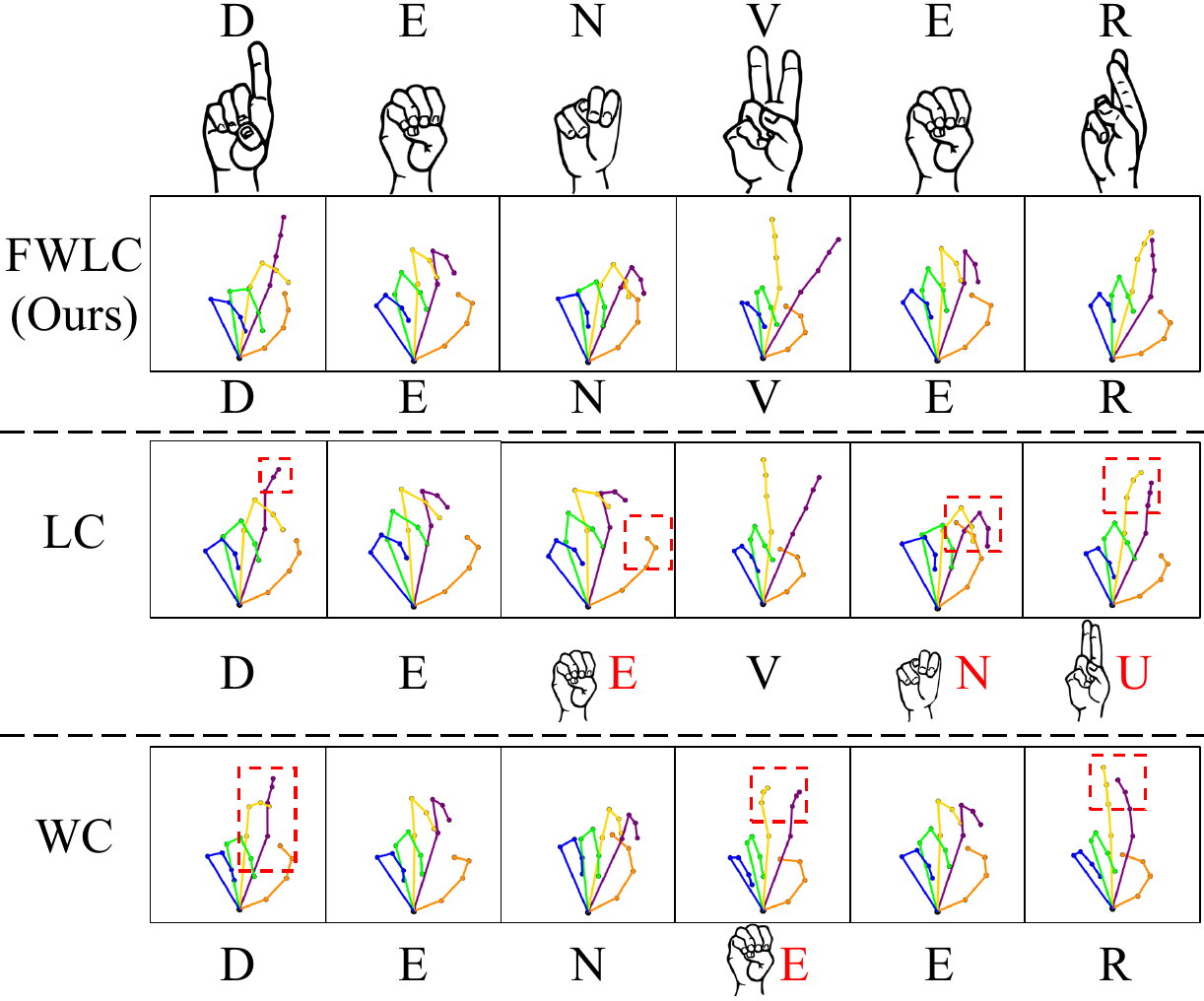}
    \vspace{-7mm}
    \caption{Generated fingerspelling pose sequences for the prompt ``denver" under FWLC (frame-wise letter-conditioned), LC (letter-conditioned), and WC (word-conditioned)~\cite{radford2021learning}. The prediction results from our recognizer are shown below each generated sequence. Incorrect poses are highlighted by red dashed rectangles, while incorrect predictions are shown in red.}
    \label{fig:generated_motion}
    \vspace{-3mm}
\end{figure}

Tab.~\ref{tab:ablation_generator_condition} shows that the fingerspelling pose sequences generated with the FWLC strategy are the most interpretable to each recognizer, and all recognizers exhibit a consistent preference in the order of FWLC, LC, and WC. As illustrated in Fig.~\ref{fig:generated_motion}, LC fails to preserve the letter order within a sequence, leading to disordered poses for letters \textit{n} and \textit{e}, and produces inaccurate poses for \textit{r}. In contrast, WC lacks letter-level feature encoding, resulting in oversmoothed and less distinctive motions, particularly for letters \textit{d} and \textit{v}, where the index finger fails to extend fully. Our recognizer predicts the letters as ``denver" (Acc.: 100), ``devnu" (Acc.: 50.0), and ``dener" (Acc.: 83.3) for FWLC, LC, and WC, respectively. Detailed pipelines for different types of conditioning, along with quantitative evaluations of generation quality, are provided in the supplementary material.

\begin{table}[t]
    \centering
    \caption{Evaluation of different conditioning strategies in the generator. 
    The results show the letter accuracy of each recognizer on the generated fingerspelling pose sequences based on the unique test-set words in CFSW~\cite{shi2018american} and CFSWP~\cite{shi2019fingerspelling}. WC, LC, and FWLC denote word-conditioned, letter-conditioned, and frame-wise letter-conditioned, respectively.
    }
    \vspace{-2mm}
    \begin{tabular}{l|c|c}
    \hline
    Generator condition & PoseNet~\cite{fayyazsanavi2024fingerspelling} & Our recognizer \\
    \hline
    WC~\cite{radford2021learning} & 19.9 & 23.3 \\
    LC & 26.4 & 40.2 \\
    FWLC (Ours) & \textbf{63.5} & \textbf{82.3} \\
    \hline
    \end{tabular}
    \label{tab:ablation_generator_condition}
\end{table}

\begin{table}[t]
    \centering
    \caption{Latency, throughput and letters per second comparison measured on 868 samples from CFSW~\cite{shi2018american}. BS denotes batch size.}
    \vspace{-2mm}
    \setlength{\tabcolsep}{4pt}
    \begin{tabular}{l|c|rrrr}
    \hline
    Method & BS & $t_{lat}$ ($\downarrow$) & $R_{tp}$ ($\uparrow$) & $R_{lps}$ ($\uparrow$) & FPS ($\uparrow$) \\
    \hline
    PoseNet~\cite{fayyazsanavi2024fingerspelling} & 1 & 4,282 & 0.2 & 1 & 6 \\
    Ours & 1 & \textbf{39} & \textbf{22.0} & \textbf{106} & \textbf{962} \\
    \hline
    Ours & 32 & \textbf{6} & \textbf{149.8} & \textbf{725} & \textbf{6,356} \\
    \hline
    \end{tabular}
    \label{tab:latency}
    \vspace{-3mm}
\end{table}

\subsection{Inference Speed Comparison}
In Tab.~\ref{tab:latency}, we compare the inference speed of our recognizer with that of PoseNet~\cite{fayyazsanavi2024fingerspelling} on an A40 GPU. Our model achieves nearly 100 times faster processing (962 FPS vs. 6 FPS), clearly demonstrating capability for real-time recognition. This substantial reduction in latency primarily stems from removing post-processing steps such as re-ranking~\cite{fayyazsanavi2024fingerspelling}, as our recognizer effectively handles inputs through its architectural design and auxiliary losses that promote robust and accurate predictions without additional refinement. Furthermore, the inference speed can be further improved through batch processing.

\section{Conclusion}
\label{sec:conclusion}
In this paper, we have presented OpenFS, offered as an open-source contribution to the research community.
A multi-hand-capable fingerspelling recognizer processes hand pose sequences with a dual-level positional encoding and enhances cross-attention through a signing-hand focus loss and a monotonic alignment loss.
It enables implicit detection of the signing hand and learning discriminative pose representations, leading to significant improvements in both signing-hand detection and letter accuracy, compared to previous methods relying on the explicit signing-hand detection and the CTC loss.
Furthermore, our recognizer requires no post-processing, enabling real-time inference.
In addition, unlike previous text-to-motion works that operate only at the word level and ignore the underlying letter-level structure, our approach incorporates a coarse-to-fine frame-wise letter annotation method and a frame-wise letter-conditioned generator, thereby providing a more intuitive and effective pipeline for the generation of fingerspelling pose sequences.
Using the frame-wise letter-conditioned generator, fingerspelling pose sequences for arbitrary words can be produced for both training and evaluation without human labor.
With the aid of this generator, we construct FSNeo to evaluate the generalizability of models to out-of-vocabulary words.
We also enhance recognition performance by synthesizing additional training data.

\noindent\textbf{Acknowledgment.} This work was supported by Institute of Information and communications Technology Planning and evaluation (IITP) grant funded by the Korea government (MSIT) (RS-2022-II220010, Development of Korean sign language translation service technology for the deaf in medical environment, 50\%, and RS-2022-II220290,  Visual Intelligence for SpaceTime Understanding and Generation based on Multilayered Visual Common Sense, 50\%).

{
    \small
    \bibliographystyle{ieeenat_fullname}
    \bibliography{main}
}

\clearpage
\appendix
\twocolumn[
\begin{center}

\vspace{1.5em}

{\Large \bfseries
OpenFS: Multi-Hand-Capable Fingerspelling Recognition with Implicit Signing-Hand Detection and Frame-Wise Letter-Conditioned Synthesis\\
\vspace{0.75em}
- Supplementary -}

\vspace{0.75em}

{\large
Junuk Cha$^1$ \qquad
Jihyeon Kim$^2$ \qquad
Han-Mu Park$^3$
\par}

\vspace{0.25em}

{\normalsize
$^1$KAIST \quad
$^2$KT \quad
$^3$KETI
\par}

\vspace{1.5em}

\end{center}
]

\renewcommand{\thefigure}{S\arabic{figure}}
\renewcommand{\thetable}{S\arabic{table}}
\renewcommand{\thesection}{S\arabic{section}}
\renewcommand{\theequation}{S\arabic{equation}}
\newcommand{\refcolor}[1]{{\color{cvprblue}{#1}}}

In the supplementary material, we provide additional explanations, extended results, and a discussion of limitations and future work, organized as the table of contents below.

\vspace{2mm}
\noindent\rule{\linewidth}{0.6pt}
\vspace{1mm}

\begin{itemize}
    \item \textbf{Additional Explanations}
    \begin{itemize}
        \item \hyperref[sec:pose_preprocess]{S1. Pose Extraction and Preprocessing}
        \item \hyperref[sec:loss]{S2. Loss Function Details}
        \begin{itemize}
            \item \hyperref[sec:ablation_loss]{S2-1. Ablation Study on SF and MA Losses}
        \end{itemize}
        \item \hyperref[sec:annotation]{S3. Coarse Frame-Wise Letter Annotation}
        \item \hyperref[sec:implementation]{S4. Implementation Details}
    \end{itemize}

    \item \textbf{Additional Experiments and Results}
    \begin{itemize}
        \item \hyperref[sec:pose_representation]{S5. Comparison of Input Pose Representations}
        \item \hyperref[sec:conditioning]{S6. Comparison of Conditioning Strategies for Generator}
        \item \hyperref[sec:model_efficiency]{S7. Model Efficiency}
        \item \hyperref[sec:error_sensitivity]{S8. Stabilized Error-Type Sensitivity}
        \item \hyperref[sec:long_words]{S9. Improving Robustness to Long Words}
        \item \hyperref[sec:additional_metric]{S10. Evaluation on Additional Metric}
        \item \hyperref[sec:fsboard]{S11. Evaluation on FSBoard}
        \item \hyperref[sec:qualitative]{S12. More Qualitative Results}
    \end{itemize}

    \item \textbf{Limitations and Future Directions}
    \begin{itemize}
        \item \hyperref[sec:failure_cases]{S13. Failure Case of Implicit Signing-Hand Detection}
        \item \hyperref[sec:pose_noise]{S14. Sensitivity Analysis under Pose Noise}
        \item \hyperref[sec:limitations]{S15. Limitations and Future Work}
    \end{itemize}
\end{itemize}

\vspace{2mm}
\noindent\rule{\linewidth}{0.6pt}
\vspace{1mm}
\section{Pose Extraction and Preprocessing}
\label{sec:pose_preprocess}

Following PoseNet~\cite{fayyazsanavi2024fingerspelling}, we employ the MediaPipe Holistic framework~\cite{lugaresi2019mediapipe} to extract multi-hand pose sequences from RGB video frames, where each hand pose is represented by 21 joints in 2D space. We normalize each pose by translating it so that the origin is set to the midpoint between the minimum and maximum coordinates of all joints. The translated coordinates are then divided by the maximum absolute coordinate value and multiplied by 0.5, resulting in values scaled to the range $[-0.5, 0.5]$. We construct the input multi-hand pose sequence by concatenating the normalized pose sequences of multiple hands along the temporal dimension.

In multi-person scenarios, since MediaPipe~\cite{lugaresi2019mediapipe} does not support multi-person hand pose extraction, we employ YOLOv11~\cite{yolo11_ultralytics} to detect all individuals appearing in a video. For frames containing multiple people, each detected person is cropped and processed independently using MediaPipe to extract hand poses. The extracted hand poses are then normalized and concatenated along the temporal dimension following the same procedure described above.

In addition, MediaPipe~\cite{lugaresi2019mediapipe} provides left–right hand identity, while YOLOv11~\cite{yolo11_ultralytics} provides person identity; we combine these two signals into a unified hand-identity encoding, which is used in both our dual-level positional encoding and the Signing-Hand Focus loss. This unified hand-identity encoding enables our multi-hand-capable recognizer to robustly process hand poses from multiple hands and multiple people simultaneously.
\section{Loss Function Details}
\label{sec:loss}

\noindent\textbf{Signing-hand focus loss for $N$ hands.}
While the main paper explains the signing-hand focus loss using the two-hand case (right and left hands) for clarity, the formulation naturally extends to an arbitrary number of hands. In the general setting, the decoder cross-attention tensor is defined as:
$\mathbf{A} \in \mathbb{R}^{L_d \times |W| \times T}$,
where $L_d$ is the number of decoder layers, 
$W = \{W_i\}_{i=1}^{|W|}$ is the output letter-token sequence of length $|W|$,
and $T = \sum_{n=1}^{N} T_n$ is the total number of pose tokens, 
with $T_n$ denoting the number of pose tokens for the $n$-th hand and $N$ denoting the total number of hands. 
The layer-averaged cross-attention is computed as:
\begin{eqnarray}
    \widetilde{\mathbf{A}}_{W_i, t}
    = \frac{1}{L_d} \sum_{\ell=1}^{L_d} \mathbf{A}_{\ell, W_i, t}
    ,\qquad 
    \widetilde{\mathbf{A}} \in \mathbb{R}^{|W| \times T}.
\end{eqnarray}

Each pose token is associated with a hand-identity label 
$h \in \{1,~\dots,~N\}$, and these labels form a one-hot matrix:
\begin{eqnarray}
    \mathbf{H} \in \mathbb{R}^{T \times N}, \qquad
    H_{t,h} = 
        \begin{cases}
        1 & 
          \begin{aligned}
          &\text{if the pose token at position } t \\
          &\text{belongs to hand } h,
          \end{aligned} \\[3pt]
        0 & \text{otherwise}.
        \end{cases}
\end{eqnarray}

The attention contribution from hand $h$ to letter token $W_i$ is computed as:
\begin{eqnarray}
    a_{W_i,h} = \sum_{t=1}^{T} \widetilde{A}_{W_i, t}\, H_{t,h},
    \qquad
    \sum_{h=1}^{N} a_{W_i,h} = 1.
\end{eqnarray}

To encourage each letter token $W_i$ to place its attention on a single signing hand,
the entropy of the hand-attention distribution is minimized:
\begin{eqnarray}
    \mathcal{E}_{W_i,h}
    = - a_{W_i,h} \log \left( a_{W_i,h} + \epsilon \right).    
\end{eqnarray}

The signing-hand focus loss is defined as:
\begin{eqnarray}
    \mathcal{L}_{\text{SF}}
    = \frac{1}{|W|} \frac{1}{N} 
    \sum_{i=1}^{|W|}
    \sum_{h=1}^{N} 
    \mathcal{E}_{W_i,h}.    
\end{eqnarray}

\noindent\textbf{Monotonic alignment loss.}
To enforce a monotonic correspondence between the pose-token sequence and the letter sequence, the cross-attention tensor 
$\mathbf{A} \in \mathbb{R}^{L_d \times |W| \times T}$ 
is accumulated along the temporal dimension.
For each letter token $W_i$, the cumulative attention is computed as:
\begin{eqnarray}
\mathbf{C}_{\ell, W_i, t} &=& \sum_{t'=1}^{t} \mathbf{A}_{\ell, W_i, t'},
\end{eqnarray}

The temporal change between $W_{i-1}$ and $W_i$ (for $i \ge 2$) is:
\begin{eqnarray}
\Delta_{\ell, W_i, t} &=& \mathbf{C}_{\ell, W_i, t} - \mathbf{C}_{\ell, W_{i-1}, t}.
\end{eqnarray}

A monotonicity violation occurs when $W_i$ places more attention on earlier pose-token positions than $W_{i-1}$. 
Such violations are defined as:
\begin{eqnarray}
\mathbf{V}_{\ell, W_i, t} &=& \max\!\left(\Delta_{\ell, W_i, t},\, 0\right).
\end{eqnarray}

The final monotonic alignment loss is:
\begin{eqnarray}
\mathcal{L}_{\text{MA}}
&=& \frac{1}{(|W|-1)} \frac{1}{T} 
    \sum_{\ell=1}^{L_d}
    \sum_{i=2}^{|W|}
    \sum_{t=1}^{T}
    \mathbf{V}_{\ell, W_i, t}.
\end{eqnarray}

\subsection{Ablation Study on SF and MA Losses}
\label{sec:ablation_loss}
To analyze the individual contributions of the signing-hand focus (SF) loss and the monotonic alignment (MA) loss, we conduct ablation experiments by selectively removing each component from the full objective. Without either auxiliary loss, the baseline model achieves 74.8 letter accuracy. Applying only the SF loss or only the MA loss results in 74.7 letter accuracy in both cases. When both losses are jointly applied, performance improves to 75.4 letter accuracy. These results indicate that the SF and MA losses are complementary and are most effective when applied together.
\section{Coarse Frame-Wise Letter Annotation}
\label{sec:annotation}

To obtain frame-wise letter annotations, we leverage the recognizer's decoder cross-attention, which encodes the correspondence between output letter tokens and input pose frames. For each decoder layer $\ell \in \{1,\dots,L_d\}$, let $\mathbf{A}_{\ell} \in \mathbb{R}^{|W| \times T}$ denote the cross-attention matrix, where $|W|$ is the number of output letter tokens and $T$ is the number of input frames. 
We first compute the layer-averaged attention:
\begin{eqnarray}
    \widetilde{\mathbf{A}}_{W_i, t}
    = \frac{1}{L_d} \sum_{\ell=1}^{L_d}
      \mathbf{A}_{\ell, W_i, t},
    \qquad
    \widetilde{\mathbf{A}} \in \mathbb{R}^{|W| \times T}.
\end{eqnarray}
Each row $\widetilde{\mathbf{A}}_{W_i,:}$ represents how strongly $i$-th letter token $W_i$ attends to each frame.

\noindent\textbf{Token-wise thresholding.}
For each token $W_i$, let $\mathbf{a}_i = \widetilde{\mathbf{A}}_{W_i,:} \in \mathbb{R}^{T}$.  
We exclude the highest attention value because it frequently shows an overly 
large peak, making it an unreliable indicator of the true attention 
distribution. Using the 2nd--4th largest values instead provides a more stable 
estimate of the typical attention magnitude, and thus yields a more robust 
threshold:
\begin{eqnarray}
    \theta_i
    &=& 0.5 \cdot
        \text{mean}\!\left( \text{top-}2\text{--}4(\mathbf{a}_i) \right).
\end{eqnarray}
Frames whose attention exceeds this threshold are assigned to token $W_i$:
\begin{eqnarray}
    L_{i,t} =
    \begin{cases}
        W_i, & \text{if } a_{i,t} \ge \theta_i, \\
        -1,  & \text{otherwise}.
    \end{cases}
\end{eqnarray}
This yields a temporary label matrix
\begin{eqnarray}
    \mathbf{L} \in \mathbb{R}^{|W| \times T}.
\end{eqnarray}
An example of this matrix is shown in Fig.~4(a) of the main paper, visualized 
as a processed cross-attention map.

\noindent\textbf{Frame-level label consolidation.}
Final frame labels are obtained by collapsing $\mathbf{L}$ along the 
letter-token dimension. For each frame $t$, let
\begin{eqnarray}
    U_t = \{\, L_{i,t} \mid L_{i,t} \neq -1 \,\}.
\end{eqnarray}
If $U_t$ contains exactly one unique letter, the frame receives that label;
otherwise (when no letter is assigned or when the frame is matched by multiple 
letter tokens), it is assigned the blank symbol $\phi$:
\begin{eqnarray}
    y_t =
    \begin{cases}
        u,   & \text{if } U_t = \{u\}, \\
        \phi, & \text{otherwise}.
    \end{cases}
\end{eqnarray}
The resulting frame-wise label sequence is
\begin{eqnarray}
    \mathbf{y} = (y_1, \dots, y_T), \qquad 
    \mathbf{y} \in \{\phi\} \cup \{1,\dots,\text{char\_size}\}^{T},
\end{eqnarray}
which constitutes the coarse frame-wise letter annotation used for training the 
frame-wise annotation refiner. An example of this label sequence $\mathbf{y}$ is 
shown in Fig.~4(a) of the main paper, visualized as coarse frame-wise letter 
labels.
\clearpage
\section{Implementation Details}
\label{sec:implementation}

All experiments are implemented in PyTorch~\cite{paszke2019pytorch} and conducted on a single NVIDIA A40 GPU. Transformer~\cite{vaswani2017attention} hyperparameters, training configurations, and the embedding/output head architectures of the recognizer are summarized in \cref{tab:arch_training_transformer_recognizer,tab:arch_rec_embed_and_head}, while those of the generator are provided in \cref{tab:arch_training_transformer_generator,tab:arch_gen_embed_and_head}. The architecture of the frame-wise annotation refiner are described in Tab.~\ref{tab:arch_anno_refiner}. Following PoseNet~\cite{fayyazsanavi2024fingerspelling}, the character set includes 26 lowercase English letters, several special symbols such as the space character, and two additional tokens, \texttt{<start>} and \texttt{<end>}, resulting in 33 characters in total.

\begin{table}[h]
\centering
\caption{Implementation details of the Transformer and training configurations for the recognizer.}
\setlength{\tabcolsep}{4pt}
\begin{tabular}{lc}
\hline
Component & Recognizer \\
\hline
Architecture & Transformer encoder--decoder \\
Layers & $L_\text{enc}/L_\text{dec} = 3/3$ \\
Hidden dimension & 256 \\
Feed-forward dimension & 2,048 \\
Attention heads & 8 \\
Activation & GELU \\
Dropout & 0.1 \\
\hline
Positional encoding & Dual-level (hand+time) \\
Optimizer & Adam~\cite{kingma2015adam} \\
Learning rate & $1\mathrm{e}{-4}$ \\
Epochs & 20 \\
LR schedule & decayed by 0.1 every 10 epochs \\
Batch size & 64 \\
\hline
\end{tabular}
\label{tab:arch_training_transformer_recognizer}
\end{table}

\begin{table}[h]
\centering
\caption{Architectures of the encoder-side pose embedding, decoder-side character embedding, and decoder head used in the \textbf{recognizer}. 
$\texttt{pose\_dim}$ corresponds to $21~(\text{joints}) \times 2~(x, y) = 42$. We set \texttt{char\_size} = 33, and the additional index is reserved for padding.}
\setlength{\tabcolsep}{4pt}
\begin{tabular}{lcc}
\hline
Layer & Input Dimension & Output Dimension \\
\hline
\multicolumn{3}{l}{\textbf{Encoder: Pose embedding}} \\
Linear & $\texttt{pose\_dim}$ & $256$ \\
LayerNorm & $256$ & $256$ \\
ReLU & $256$ & $256$ \\
Linear & $256$ & $256$ \\
LayerNorm & $256$ & $256$ \\
ReLU & $256$ & $256$ \\
\hline
\multicolumn{3}{l}{\textbf{Decoder: Character embedding}} \\
Embedding & $\texttt{char\_size} + 1$ & $256$ \\
\hline
\multicolumn{3}{l}{\textbf{Decoder: Output head}} \\
Linear & $256$ & $128$ \\
ReLU & $128$ & $128$ \\
Linear & $128$ & $33$ \\
\hline
\end{tabular}
\label{tab:arch_rec_embed_and_head}
\end{table}

\begin{table}[h]
\centering
\caption{Implementation details of the Transformer architecture and training configurations for the generator, and the diffusion process.}
\begin{tabular}{lc}
\hline
Component & Generator (DDIM) \\
\hline
Architecture & Transformer encoder \\
Layers & $L = 8$ \\
Hidden dimension & 256 \\
Feed-forward dimension & 1,024 \\
Attention heads & 4 \\
Activation & GELU \\
Dropout & 0.1 \\
Positional encoding & Temporal \\
\hline
Optimizer & Adam~\cite{kingma2015adam} \\
Learning rate & $1\mathrm{e}{-4}$ \\
Epochs & 1,000 \\
Batch size & 20 \\
\hline
Diffusion method & DDIM~\cite{song2020denoising} \\
Noise schedule & Cosine \\
Diffusion steps & $T = 50$ \\
\hline
\end{tabular}
\label{tab:arch_training_transformer_generator}
\end{table}

\begin{table}[h]
\centering
\caption{Architectures of the embedding layers and output head used in the \textbf{generator}. 
We use \texttt{char\_size} = 33, and the additional index is reserved for padding. 
$\texttt{pose\_dim}$ corresponds to $21~(\text{joints}) \times 3~(x, y, z) = 63$.}
\setlength{\tabcolsep}{4pt}
\begin{tabular}{lcc}
\hline
Layer & Input Dimension & Output Dimension \\
\hline
\multicolumn{3}{l}{\textbf{Time embedding}} \\
Sinusoidal & 1 & 256 \\
Linear & 256 & 256 \\
SiLU & 256 & 256 \\
Linear & 256 & 256 \\
\hline
\multicolumn{3}{l}{\textbf{Pose embedding}} \\
Linear & $\texttt{pose\_dim}$ & 128 \\
\hline
\multicolumn{3}{l}{\textbf{Letter embedding}} \\
Embedding & $\texttt{char\_size} + 1$ & 128 \\
\hline
\multicolumn{3}{l}{\textbf{Output head}} \\
Linear & 256 & $\texttt{pose\_dim}$ \\
\hline
\end{tabular}
\label{tab:arch_gen_embed_and_head}
\end{table}

\begin{table}[h]
\centering
\caption{Architecture of the frame-wise annotation refiner. We set \texttt{char\_size} = 33.}
\setlength{\tabcolsep}{4pt}
\begin{tabular}{lcc}
\hline
Layer & Input Dimension & Output Dimension \\
\hline
Linear & 256 & 256 \\
LayerNorm & 256 & 256 \\
ReLU & 256 & 256 \\
Dropout ($p=0.1$) & 256 & 256 \\
Linear & 256 & $\texttt{char\_size}$ \\
\hline
\end{tabular}
\label{tab:arch_anno_refiner}
\end{table}


\clearpage
\section{Comparison of Input Pose Representations}
\label{sec:pose_representation}

In Tab.~\ref{tab:ablation_representation}, we first compare 3D and 2D input pose representations. Although Mediapipe~\cite{lugaresi2019mediapipe} provides 3D hand poses, they are substantially noisier than their 2D estimates, and this noise leads to noticeably lower recognition accuracy. A similar observation was also reported in PoseNet~\cite{fayyazsanavi2024fingerspelling}. As a result, the 2D representation yields the best performance.

We also evaluate an alternative representation that does not treat the multi-hand pose sequences as a frame-wise concatenation but instead concatenates all joint coordinates within each frame into a single vector (joint-wise concatenation). In our default setting, a pose sequence has the shape $T \times N \times J \times d$, where $T$ is the number of frames, $N$ is the number of hands, $J$ is the number of joints per hand, and $d$ is the coordinate dimension. The \textit{frame-wise} representation preserves the per-hand spatial structure, and the model effectively receives inputs of the form $(T \times N) \times J \times d$. In contrast, the \textit{joint-wise} representation collapses the $N$ hands within each frame, resulting in
$T \times (N J) \times d$, which merges informative and irrelevant hands into the same joint axis, providing a less discriminative input representation and ultimately degrading recognition performance.

\begin{table}[t]
    \centering
    \caption{Ablation on input pose representations on CFSW~\cite{shi2018american}, comparing coordinate dimension (2D vs 3D) and concatenation strategy (frame-wise vs joint-wise).}
    \begin{tabular}{cc|c}
    \hline
    Coordinate Dim. & Concatenation & Letter Accuracy \\
    \hline
    3D & frame-wise & 73.4 \\
    2D & joint-wise & 71.6 \\
    2D & frame-wise & \textbf{75.4} \\
    \hline
    \end{tabular}
    \label{tab:ablation_representation}
\end{table}

\section{Comparison of Conditioning Strategies for Generator}
\label{sec:conditioning}

\begin{table}[t]
    \centering
    \caption{FID and Diversity results across different conditioning methods. WC denotes word-conditioned generation, LC denotes letter-conditioned generation, and FWLC denotes frame-wise letter-conditioned generation. Lower FID indicates better fidelity, and Diversity results are better when the values are closer to the real data. All evaluations are repeated 20 times, and $\pm$ indicates the 95\% confidence interval.}
    \begin{tabular}{l|cc}
    \hline
    Method   & FID ($\downarrow$) & Diversity ($\rightarrow$) \\
    \hline
    Real     & - & 1.1107 ± 0.0115 \\
    \hline
    WC       & 0.4036 ± 0.0408 & 0.7803 ± 0.0015 \\
    LC       & 0.4159 ± 0.0516 & 0.8987 ± 0.0009 \\
    FWLC     & \textbf{0.3695 ± 0.0522} & \textbf{0.8996 ± 0.0006} \\
    \hline
    \end{tabular}
    \label{tab:fid_diversity_generator}
\end{table}

\begin{figure}[t]
    \centering
    \includegraphics[width=1\linewidth]{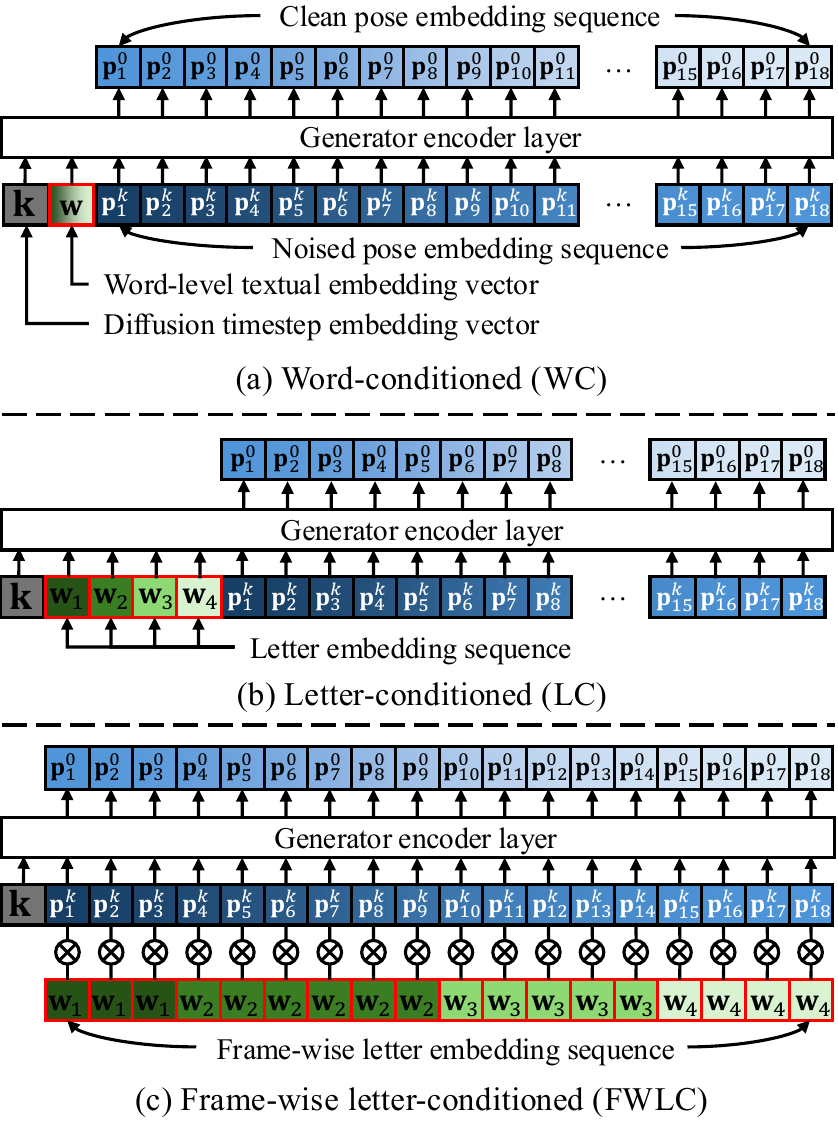}
    \vspace{-7mm}
    \caption{\textbf{Comparison of the three conditioning strategies.} 
    We omit the details of standard positional encoding~\cite{vaswani2017attention} and the embedding layers. 
    For illustration, we consider a word of length 4. 
    The symbol $\mathbf{w}$ denotes the word-level textual embedding, 
    $\mathbf{w}_i$ denotes the embedding of the $i$-th letter in the word, 
    and $\mathbf{p}_t^{k}$ denotes a pose embedding token, where $t$ is the frame index and $k$ is the diffusion timestep ($k=0$ indicates a clean pose). 
    The diffusion timestep embedding vector is represented as $\mathbf{k}$. 
    (a) WC uses a single word-level textual embedding token as input. 
    (b) LC takes a sequence of letter tokens, with one token corresponding to each character. 
    (c) FWLC aligns each pose embedding token with its corresponding letter embedding token and concatenates them in a one-to-one, frame-wise manner. 
    The different conditioning inputs are highlighted with red rectangles.
    }
    \vspace{-5mm}
\label{fig:comparison_generator}
\end{figure}

Fig.~\ref{fig:comparison_generator} shows a design comparison of the three conditioning strategies for the generator: word-conditioned (WC), letter-conditioned (LC), and frame-wise letter-conditioned (FWLC). For the WC setting, we adopt the CLIP word-level text embedding~\cite{radford2021learning} as the conditioning input following previous text-to-motion methods~\cite{guo2024momask,tevet2022motionclip,lin2023being,tevet2023human,zhang2024motiondiffuse,meng2025rethinking}. In contrast, LC and FWLC do not rely on CLIP, instead, each letter is represented by a learnable letter embedding layer that maps character indices to continuous embedding vectors. This difference causes WC to encode word-level semantics, while LC and FWLC provide letter-specific conditioning signals, which are more suitable for fingerspelling pose generation since fingerspelled hand poses are inherently defined at the letter level.


In Tab.~\ref{tab:fid_diversity_generator}, we further measure the FID and Diversity of samples produced under different conditioning strategies. Among these strategies, the frame-wise letter-conditioned (FWLC) generator delivers the best overall performance. The word-conditioned (WC) baseline conditions generation solely on a single word-level embedding, ignoring the structure of individual letters, while the letter-conditioned (LC) baseline provides letter-level cues only at coarse segment boundaries without enforcing frame-wise temporal alignment. In contrast, FWLC conditions the generator on the corresponding letter at every frame, preserving both temporal alignment and fine-grained letter-level structure. This dense conditioning leads FWLC to generate synthetic sequences with higher fidelity and more realistic diversity, achieving the strongest results across all metrics. Additional qualitative comparisons (Fig.~\refcolor{7}), along with recognizer accuracy evaluations (Tab.~\refcolor{5}) on synthetic data generated under each conditioning strategy, are provided in the main paper.


\begin{table}[t]
    \centering
    \caption{Model size, number of parameters, and inference speed (FPS) of 
    PoseNet~\cite{fayyazsanavi2024fingerspelling} and our models, grouped into the 
    multi-hand-capable (MHC) recognizer and the Frame-Wise Letter-Conditioned (FWLC) generator.}
    \setlength{\tabcolsep}{3pt}
    \begin{tabular}{l|ccc}
    \hline
    Method & Model Size (MB) & Params. (M) & FPS \\
    \hline
    PoseNet~\cite{fayyazsanavi2024fingerspelling} & 33.5 & 8.78 & 6 \\
    MHC recognizer & 33.7 & 8.81 & 962 \\
    \hline
    FWLC generator & 24.8 & 6.48 & 96 \\
    \hline
    \end{tabular}
    \label{tab:model_size}
    \vspace{-2mm}
\end{table}

\section{Model Efficiency}
\label{sec:model_efficiency}

In Table~\ref{tab:model_size}, we report the model size, parameter counts, and inference speed (FPS) of our multi-hand-capable (MHC) recognizer and Frame-Wise Letter-Conditioned (FWLC) generator, compared against PoseNet~\cite{fayyazsanavi2024fingerspelling}. Although the MHC recognizer has a similar number of parameters to PoseNet (8.81M vs.\ 8.78M), it achieves a higher inference speed (962 FPS vs.\ 6 FPS), demonstrating that our architecture is substantially more efficient while supporting multi-hand inputs. The FWLC generator is even more lightweight, with 6.48M parameters and a model size of only 24.7~MB, and operates at 96 FPS, enabling fast pose-sequence synthesis.
\begin{table}[t]
    \centering
    \caption{Comparison of deletion, substitution, and insertion error counts and their corresponding rates across different methods on CFSW~\cite{shi2018american}. The symbol $\dag$ denotes methods trained with additional synthetic data.}
    \setlength{\tabcolsep}{3pt}
    \begin{tabular}{l|ccc}
    \hline
    Method & Deletions & Substitutions & Insertions \\
    \hline
    PoseNet~\cite{fayyazsanavi2024fingerspelling} & 962(21.8\%) & 502(11.4\%) & \textbf{235(5.3\%)} \\
    Ours & \textbf{458(10.4\%)} & \textbf{390(8.8\%)} & 239(5.4\%) \\
    \hline
    PoseNet$^\dagger$~\cite{fayyazsanavi2024fingerspelling} & 850(19.2\%) & \textbf{339(7.7\%)} & \textbf{173(3.9\%)} \\
    Ours$^\dagger$ & \textbf{336(7.6\%)} & 344(7.8\%) & 304(6.9\%) \\
    \hline
    \end{tabular}
    \label{tab:error_count}
    \vspace{-5mm}
\end{table}

\section{Stabilized Error-Type Sensitivity}
\label{sec:error_sensitivity}

In Tab.~\ref{tab:error_count}, Ours$^\dagger$ (the symbol $\dagger$ denotes methods trained with additional synthetic data) shows superior performance in the case of \emph{Deletions} and comparable results in \emph{Substitutions} and \emph{Insertions}, while maintaining balanced error rates across all error types. Compared to Ours, Ours$^\dagger$ exhibits a slight increase in insertion errors, but achieves a substantial reduction in deletion errors and a modest reduction in substitution errors. Notably, PoseNet and PoseNet$^\dagger$~\cite{fayyazsanavi2024fingerspelling} exhibits an extreme imbalance between deletion and insertion errors; deletions occur 4–5 times more frequently than insertions. This indicates that the model has a strongly conservative decoding behavior, often choosing not to output a letter under uncertainty.
\section{Improving Robustness to Long Words}
\label{sec:long_words}

As shown in Tab.~\ref{tab:word_length}, both recognizers exhibit clear performance degradation as word length increases when trained only on the original dataset (PoseNet~\cite{fayyazsanavi2024fingerspelling}: 65.6$\rightarrow$59.4; Ours: 79.2$\rightarrow$73.4). However, incorporating the synthetic data generated by our frame-wise letter-conditioned generator substantially reduces this gap. For PoseNet$^\dagger$~\cite{fayyazsanavi2024fingerspelling}, the short-to-long difference shrinks from 6.2 to 0.4, effectively eliminating the length-related drop. Similarly, Ours$^\dagger$ reduces the gap from 5.8 to 3.4, demonstrating improved robustness on longer words.

\begin{table}[t]
    \centering
    \caption{Letter accuracy across word-length bins (1–4 and 5+) on CFSW~\cite{shi2018american}. Diff. denotes the accuracy drop between short (1–4) and long (5+) words. Both PoseNet~\cite{fayyazsanavi2024fingerspelling} and our model show accuracy drops on longer words, while their synthetic-data variants$^\dagger$ markedly reduce this gap.}
    \begin{tabular}{l|cc|c}
    \hline
    Method & 1-4 & 5+ & Diff. \\
    \hline
    PoseNet~\cite{fayyazsanavi2024fingerspelling} & 65.6 & 59.4 & 6.2 \\
    Ours & \textbf{79.2} & \textbf{73.4} & \textbf{5.8} \\
    \hline
    PoseNet$^\dagger$~\cite{fayyazsanavi2024fingerspelling} & 69.2 & 68.8 & \textbf{0.4} \\
    Ours$^\dagger$ & \textbf{80.0} & \textbf{76.6} & 3.4 \\
    \hline
    \end{tabular}
    \label{tab:word_length}
    \vspace{-2mm}
\end{table}
\section{Evaluation on Additional Metric}
\label{sec:additional_metric}

\begin{table}[t]
    \centering
    \caption{We compare PoseNet~\cite{fayyazsanavi2024fingerspelling}, Ours, and their variants$^\dagger$ trained with additional synthetic data on CFSW~\cite{shi2018american}, CFSWP~\cite{shi2019fingerspelling} and FSNeo, using top-1 accuracy as the evaluation metric. Parenthesized numbers indicate the accuracy improvements achieved by using the additional synthetic data$^\dagger$.}
    \begin{tabular}{l|ccc}
    \hline
    Method & CFSW & CFSWP & FSNeo \\
    \hline
    PoseNet~\cite{fayyazsanavi2024fingerspelling} & 28.8 & 25.7 & 4.7 \\
    Ours & \textbf{44.6} & \textbf{40.4} & \textbf{24.1} \\
    \hline
    PoseNet$^\dagger$~\cite{fayyazsanavi2024fingerspelling} & 36.9(\textbf{+8.1}) & 36.0(\textbf{+10.3}) & 64.2(\textbf{+59.5}) \\
    Ours$^\dagger$ & \textbf{50.5}(+5.9) & \textbf{48.5}(+8.1) & \textbf{82.0}(+57.9) \\
    \hline
    \end{tabular}
    \label{tab:additional_metric}
    \vspace{-5mm}
\end{table}

As shown in Tab.~\ref{tab:additional_metric}, we utilize top-1 accuracy as an additional evaluation metric to complement the letter-accuracy results reported in the main paper. Letter accuracy evaluates character-level correctness by accounting for substitution, deletion, and insertion errors, whereas top-1 accuracy is a word-level metric that counts a prediction as correct only when the entire word is recognized exactly. Under this more challenging metric, our recognizer achieves significantly higher performance than PoseNet~\cite{fayyazsanavi2024fingerspelling} on all datasets. Furthermore, the variants trained with our additional synthetic data$^\dagger$, generated by the frame-wise letter-conditioned (FWLC) generator, show substantial improvements in top-1 accuracy. This indicates that FWLC produces synthetic sequences that are not only realistic at the frame level but also highly effective for enhancing overall word-level recognition performance.
\begin{table}[t]
    \centering
    \caption{On the FSboard dataset~\cite{georg2025fsboard}, our multi-hand-capable recognizer substantially outperforms ByT5-based methods~\cite{georg2025fsboard,xue2022byt5} while using 34$\times$ fewer parameters. ByT5-s and ByT5-p denote the ByT5 model trained from scratch and the pretrained variant, respectively.
    }
    \setlength{\tabcolsep}{3pt}
    \begin{tabular}{l|ccc}
    \hline
    Method     & Letter Acc. & Top-1 Acc. &  Params. (M) \\
    \hline
    ByT5-s~\cite{georg2025fsboard,xue2022byt5} & 66.2 & 17.9 & 300M \\
    ByT5-p~\cite{georg2025fsboard,xue2022byt5} & 88.9 & 52.9 & 300M \\
    Ours               & \textbf{93.7} & \textbf{59.4} & 8.81M \\
    \hline
    \end{tabular}
    \label{tab:evaluation_on_FSboard}
    \vspace{-5mm}
\end{table}

\section{Evaluation on FSboard}
\label{sec:fsboard}

We evaluate our multi-hand-capable recognizer on FSboard~\cite{georg2025fsboard}, a large-scale ASL fingerspelling dataset collected from smartphone recordings.
FSboard provides 151K sequences (train: 126K, validation: 12K, test: 13K) totaling 3.2M characters across 147 signers. Unlike existing ASL fingerspelling datasets, FSboard includes diverse content categories such as addresses, URLs, names, and phone numbers, and contains not only alphabetic fingerspelling but also digits and special characters, offering a significantly broader label space.

As shown in Tab.~\ref{tab:evaluation_on_FSboard}, the ByT5~\cite{xue2022byt5}-based method trained on FSboard~\cite{georg2025fsboard} from scratch (ByT5-s) achieves 66.2 letter accuracy and 17.9 top-1 accuracy, while the pretrained variant (ByT5-p) achieves 88.9 letter accuracy and 52.9 top-1 accuracy. In contrast, our recognizer achieves 93.7 letter accuracy and 59.4 top-1 accuracy, while using 34 times fewer parameters (8.81M vs 300M), demonstrating substantially higher efficiency and effectiveness. Notably, FSboard contains not only alphabetic fingerspelling but also digits and special characters. Some of these symbols share similar hand poses across different character domains (e.g., letters vs. numbers), which introduces additional ambiguity. Despite this challenge, our model remains robust to such cross-domain pose similarity and consistently outperforms the ByT5-based methods. Moreover, while the ByT5-pretrained method benefits from large-scale pretraining on massive text data, our recognizer is trained without such pretraining and still achieves superior accuracy with significantly fewer parameters, highlighting the effectiveness of our architecture.

\section{More Qualitative Results}
\label{sec:qualitative}


In ~\cref{fig:comparison_results_1,fig:comparison_results_2,fig:comparison_results_3}, we present the qualitative recognition results of PoseNet~\cite{fayyazsanavi2024fingerspelling}, PoseNet$^\dagger$, Ours, and Ours$^\dagger$ on ChicagoFSWild~\cite{shi2018american}. The symbol $\dagger$ denotes models trained with additional synthetic data generated by our frame-wise letter-conditioned generator. We also report the letter accuracy for each prediction. Letter accuracy is the standard metric defined as the proportion of correctly predicted letters after accounting for substitutions, deletions, and insertions.

In addition, Fig.~\ref{fig:labeling} illustrates qualitative examples of our coarse-to-fine frame-wise letter annotation process, highlighting the alignment between refined labels and human annotations.


\begin{figure}[t]
    \centering
    \includegraphics[width=1\linewidth]{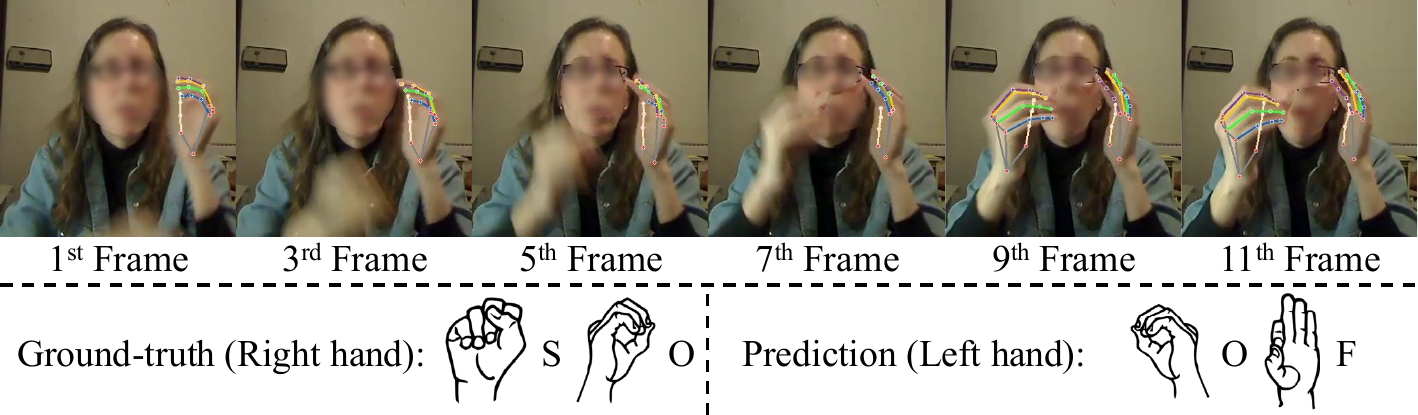}
    \vspace{-7mm}
    \caption{\textbf{The only failure case of implicit signing-hand detection.} Although the signer uses the right hand for fingerspelling, the short video and motion blur in early frames cause the model to incorrectly identify the left hand as the signing hand. The ground-truth is ``so" with the right hand, whereas the prediction is ``of" with the left hand.}
    \label{fig:detection_failure}
    \vspace{-4mm}
\end{figure}

\section{Failure Case of Implicit Signing-Hand Detection}
\label{sec:failure_cases}

While our implicit signing-hand detection module performs reliably in most cases, we identify a single failure case in the test set. In this example, the signing-hand pose appears only in the last four frames out of twelve after the motion blur subsides, as shown in Fig.~\ref{fig:detection_failure}. Under such severe blur, even humans may struggle to correctly determine the signing hand.

\begin{figure}[t]
    \centering
    \includegraphics[width=1\linewidth]{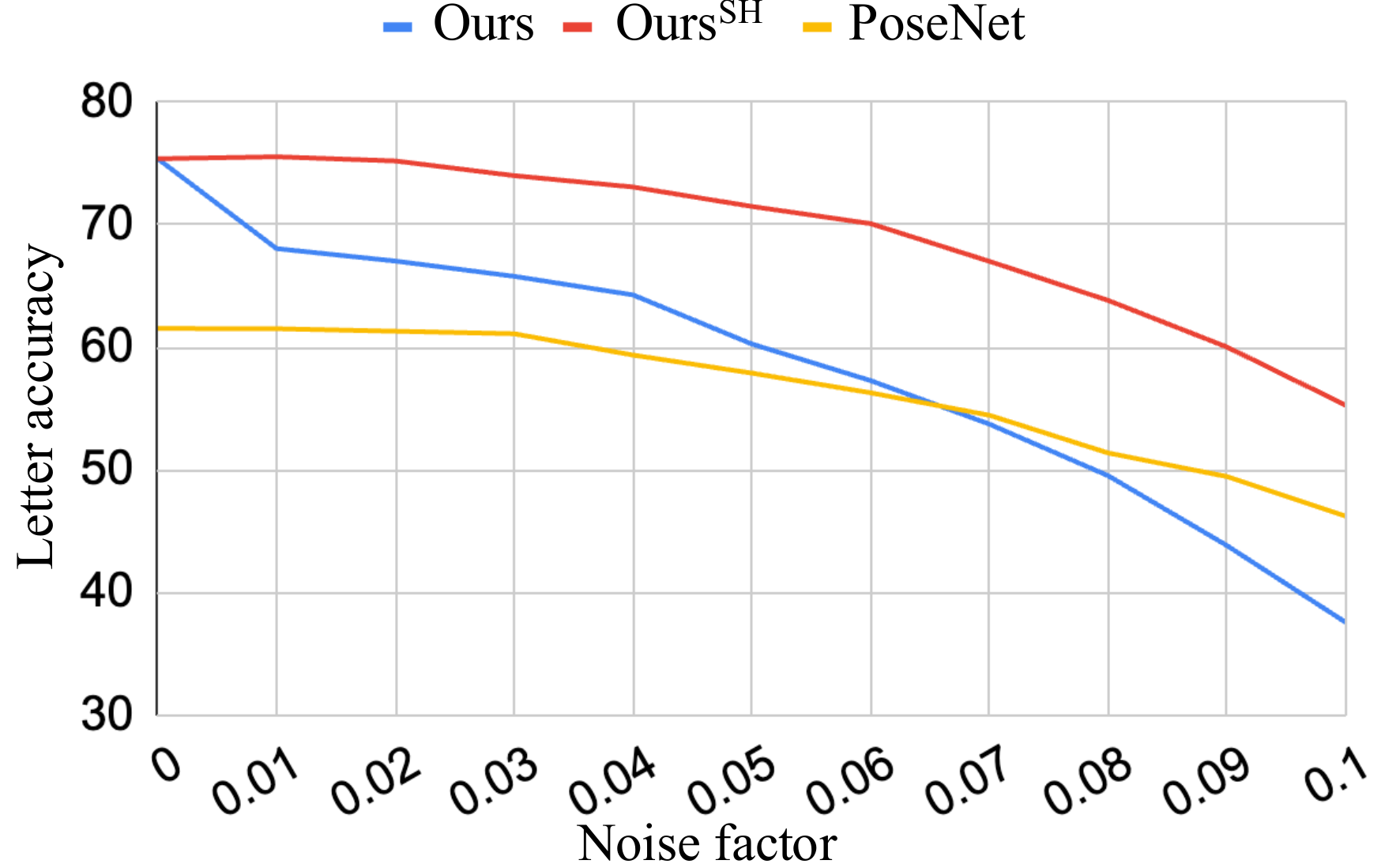}
    \vspace{-5mm}
    \caption{
    Robustness to pose noise. We increment the noise factor by 0.01 at each step. Ours$^{\text{SH}}$ refers to the variant that first identifies the signing hand via cross-attention and then performs recognition using only the selected hand. This single-hand variant maintains substantially higher robustness than both our full model (Ours) and PoseNet across all noise factors.
    }
    \label{fig:noised_pose}
    \vspace{-5mm}
\end{figure}

\section{Sensitivity Analysis under Pose Noise}
\label{sec:pose_noise}

Fig.~\ref{fig:noised_pose} shows the robustness comparison under pose noise among PoseNet~\cite{fayyazsanavi2024fingerspelling}, Ours, and Ours$^\text{SH}$. Ours$^\text{SH}$ denotes the variant of our model that first identifies the signing hand via cross-attention and then performs recognition using only the selected hand. We add Gaussian noise to the original pose $P$ as $\tilde{P} = P + \sigma_{\text{noise}} \, \mathcal{N}(0, I)$, where the noise factor $\sigma_\text{noise}$ increases from 0 to 0.10 in increments of 0.01 (each step corresponding to 1\% of the pose value range $[-0.5, 0.5]$).

Ours exhibits clear vulnerability to pose noise. Its performance drops sharply even at $\sigma_\text{noise}=0.01$, and it falls below PoseNet starting from $\sigma_\text{noise}=0.07$. This degradation occurs because the non-signing hand introduces additional noise that interferes with correct recognition. To validate this hypothesis, we evaluate a single-hand variant. Ours$^\text{SH}$ shows substantially better robustness to pose noise than both Ours and PoseNet across all noise levels.
\section{Limitations and Future Work}
\label{sec:limitations}


The publicly released dataset exposes multiple challenges: annotation-pose mismatch and articulation-induced label distortion.

\noindent\textbf{Annotation–Pose Mismatch.} As shown in Fig.~\ref{fig:detection_failure}, the ground-truth label for this sample is ``so", but the actual hand configuration resembles ``o". This type of mismatch arises from manual annotation ambiguity and highlights the difficulty of accurate labeling when frames are noisy or blurred. Such mismatches degrade recognizer performance when present in the training data, and when they appear in the test set, they impose an upper bound on achievable accuracy.

\noindent\textbf{Articulation-Induced Label Distortion.} The dataset also contains cases where the signer does not fully articulate certain letters. The annotations faithfully follow the visible evidence, yet the resulting labels may form non-existent words. For instance, ``ASL Companion Volume" becomes ``ASL Companion Volme" because the signer omits the letter ``u" when producing ``volume", as illustrated in Fig.~\ref{fig:comparison_results_3}. In this case, the phenomenon itself does not directly harm recognizer performance. However, depending on the annotator’s judgment, such example may be labeled as ``volume" despite the missing articulation, thereby risking a transition into an annotation–pose mismatch.


Furthermore, because our method relies on MediaPipe~\cite{lugaresi2019mediapipe} for hand-pose extraction, the resulting pose sequences inherit estimator-induced uncertainty, which may introduce additional noise for downstream models, as shown in Fig.~\ref{fig:noised_pose}. This dependency also makes the system vulnerable to jittering and motion blur, especially in low-quality or fast-moving video.

As future work, we aim to develop methods that leverage context beyond the fingerspelling clip itself. Certain failure cases such as motion blur, occlusions, incomplete articulations, and pose estimator noise remain challenging to handle within the scope of fingerspelling only. This motivates the development of a broader sign language understanding framework that incorporates surrounding signs and semantic context. By integrating fingerspelling with standard sign language, the model can reason over the full communicative context and produce more robust and reliable predictions under challenging visual conditions.

\begin{figure*}
    \centering
    \includegraphics[width=0.95\linewidth]{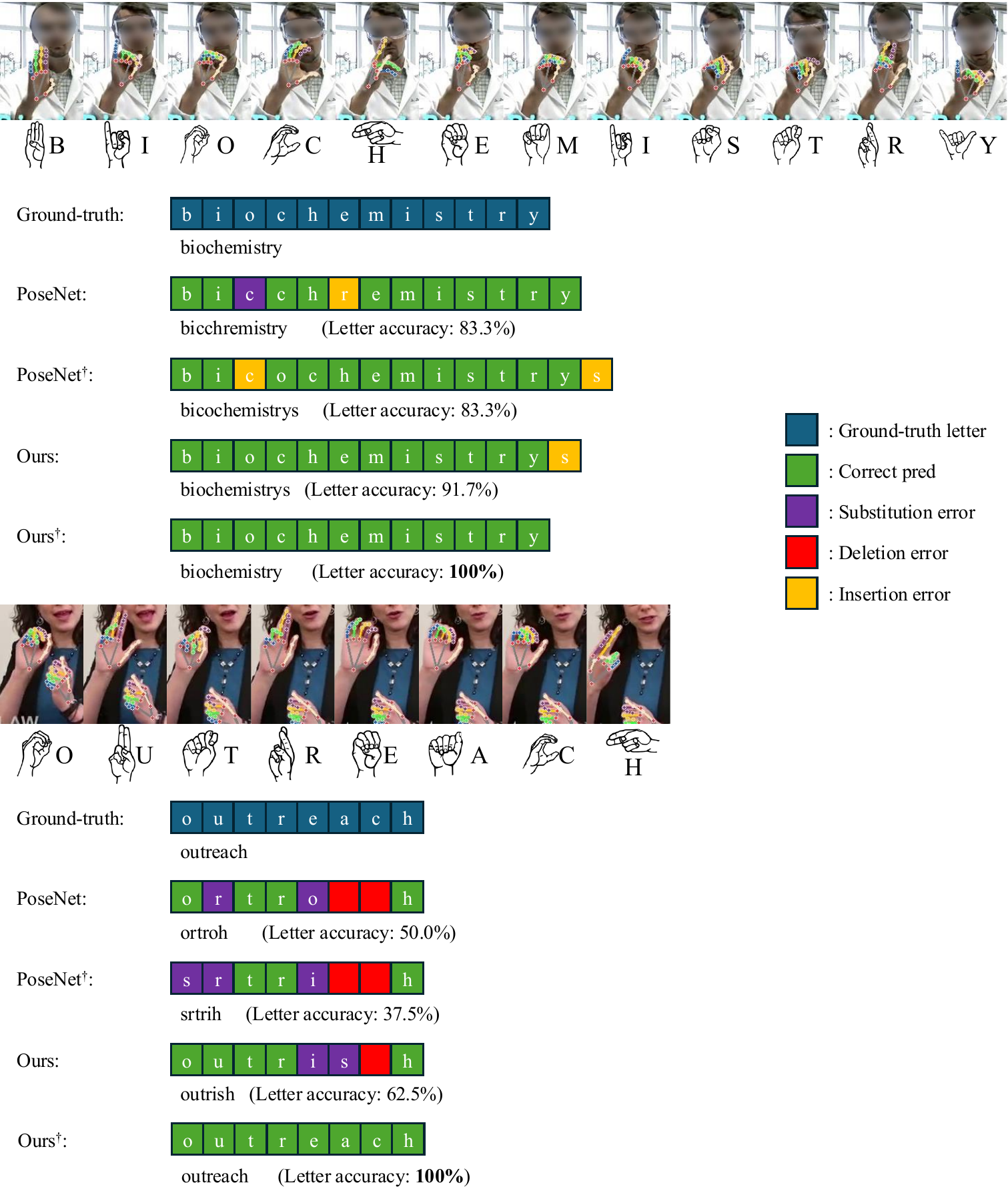}
    \caption{Qualitative recognition results on ChicagoFSWild~\cite{shi2018american}. For each example, we show the input frames, the ground-truth letters, and the predictions from PoseNet, PoseNet$^\dagger$, Ours, and Ours$^\dagger$. The symbol $\dagger$ denotes models trained with additional synthetic data generated by our frame-wise letter-conditioned generator. We also report the letter accuracy for each prediction. Colored blocks indicate different types of prediction outcomes: blue for ground-truth letters, green for correct predictions, purple for substitution errors, red for deletion errors, and yellow for insertion errors.}
    \label{fig:comparison_results_1}
\end{figure*}

\begin{figure*}
    \centering
    \includegraphics[width=0.95\linewidth]{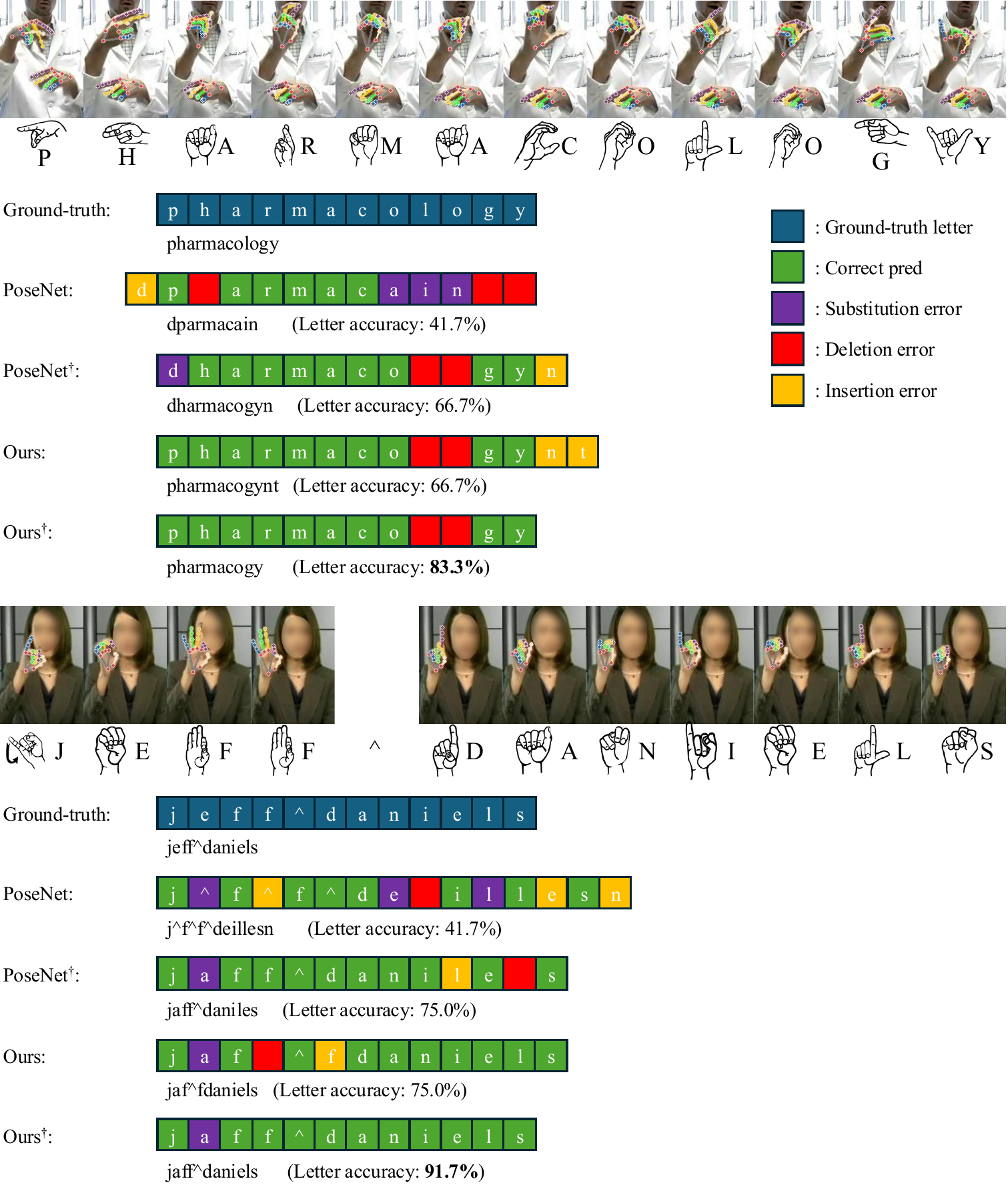}
    \caption{Qualitative recognition results on ChicagoFSWild~\cite{shi2018american}. For each example, we show the input frames, the ground-truth letters, and the predictions from PoseNet, PoseNet$^\dagger$, Ours, and Ours$^\dagger$. The symbol $\dagger$ denotes models trained with additional synthetic data generated by our frame-wise letter-conditioned generator. We also report the letter accuracy for each prediction. Colored blocks indicate different types of prediction outcomes: blue for ground-truth letters, green for correct predictions, purple for substitution errors, red for deletion errors, and yellow for insertion errors. The symbol \textasciicircum\ denotes a space character.}
    \label{fig:comparison_results_2}
\end{figure*}

\begin{figure*}
    \centering
    \includegraphics[width=0.9\linewidth]{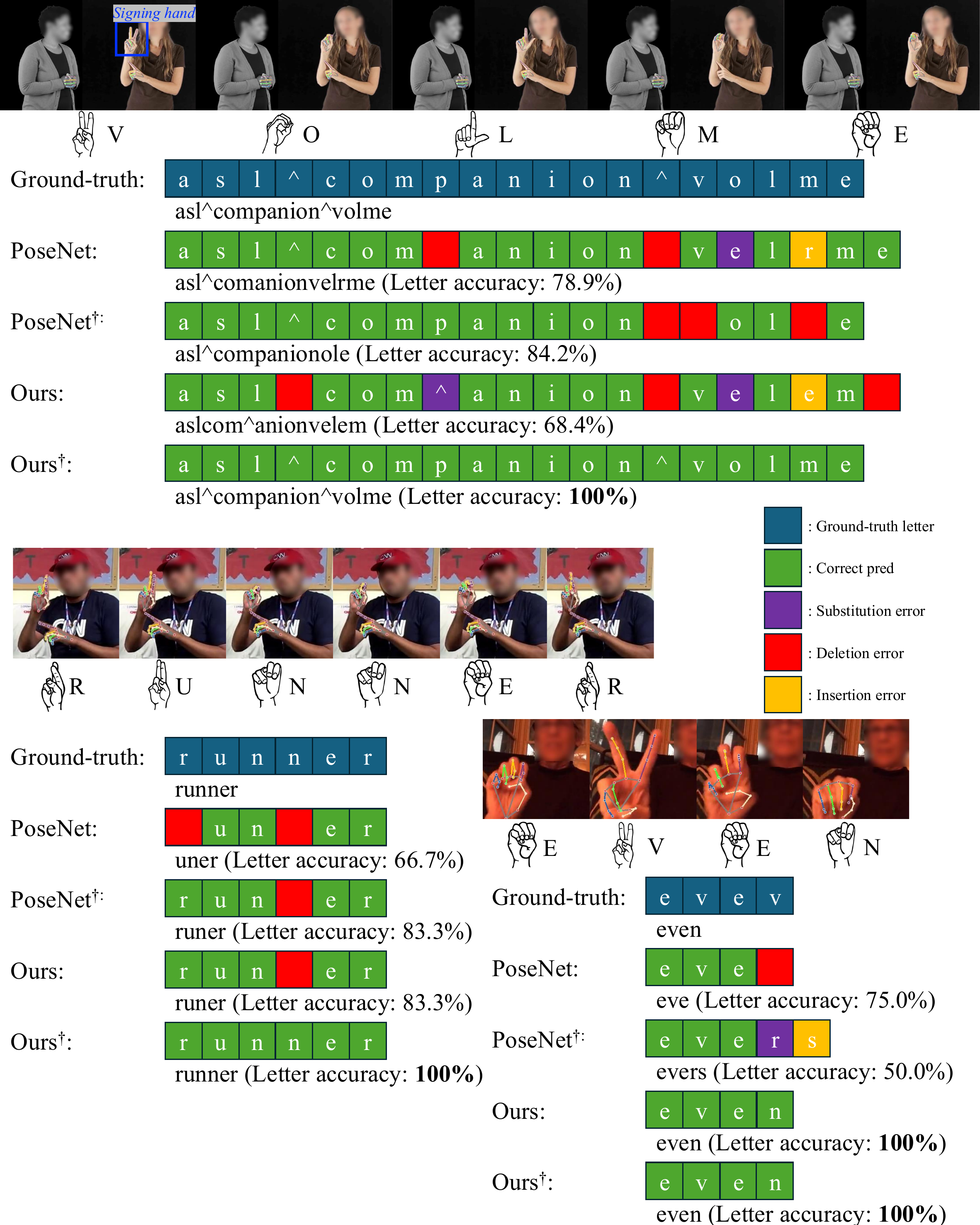}
    \caption{Qualitative recognition results on ChicagoFSWild~\cite{shi2018american}. For each example, we show the input frames, the ground-truth letters, and the predictions from PoseNet, PoseNet$^\dagger$, Ours, and Ours$^\dagger$. The symbol $\dagger$ denotes models trained with additional synthetic data generated by our frame-wise letter-conditioned generator. We also report the letter accuracy for each prediction. Colored blocks indicate different types of prediction outcomes: blue for ground-truth letters, green for correct predictions, purple for substitution errors, red for deletion errors, and yellow for insertion errors. The symbol \textasciicircum\ denotes a space character. For the first example, earlier input frames are omitted for space.}
    \label{fig:comparison_results_3}
\end{figure*}

\begin{figure*}
    \centering
    \includegraphics[width=0.9\linewidth]{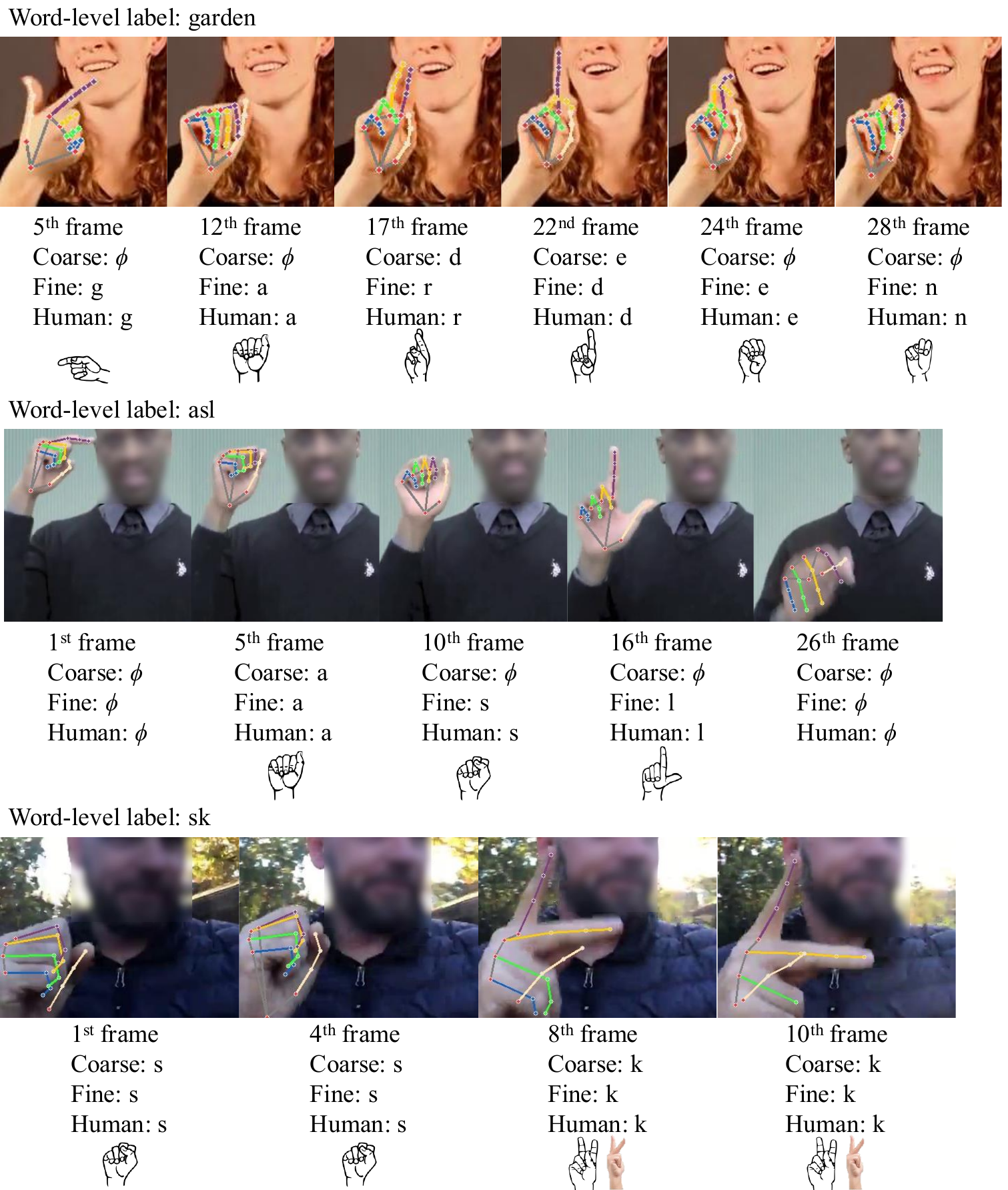}
    \caption{Qualitative examples of coarse-to-fine frame-wise letter annotation. For each word-level label (top: garden, middle: asl, bottom: sk), selected frames are shown with extracted hand keypoints. We compare coarse pseudo labels, fine-grained refined labels, and human annotations. The symbol $\phi$ denotes the absence of a letter label for the corresponding frame. The results show that the refined fine labels align more closely with human annotations, especially in frames where the coarse labels fail to assign a valid letter.}
    \label{fig:labeling}
\end{figure*}







\end{document}